 \documentclass[pmlr,twocolumn,10pt]{jmlr} 

\usepackage{booktabs}
\usepackage{siunitx}

\usepackage{bold-extra}

\usepackage{enumitem}
\usepackage{multirow}
\usepackage{amsmath}
\usepackage{amssymb}
\usepackage{mathtools}
\usepackage{pifont}

\definecolor{Green}{RGB}{0,170,85}
\definecolor{BrickRed}{RGB}{198,49,26}
\definecolor{Dandelion}{RGB}{255,185,84}

\newcommand{\cmark}{{\color{Green} \ding{51}}}
\newcommand{\xmark}{{\color{BrickRed} \ding{55}}}
\newcommand{\omark}{{\color{Dandelion} \ding{104}}}

\newcommand{\cbmark}{\ding{51}}
\newcommand{\xbmark}{\ding{55}}

\usepackage{algorithm}
\usepackage{algorithmic}

\usepackage{adjustbox}
\newcommand{\bftab}{\fontseries{b}\selectfont}
\usepackage{graphicx}
\usepackage{scalerel}
\usepackage{chngcntr}
\counterwithin{figure}{section}
\counterwithin{table}{section}
\usepackage{placeins}

\usepackage[T1]{fontenc}
\usepackage[tracking=true]{microtype}
\SetTracking{encoding=*, shape=sc}{0}

\theorembodyfont{\upshape}
\theoremheaderfont{\scshape}
\theorempostheader{:}
\theoremsep{\newline}

\jmlrvolume{225}
\jmlryear{2023}
\jmlrworkshop{Machine Learning for Health (ML4H) 2023} 

 \title[Temporal Supervised Contrastive Learning]{\vspace{-.75em}Temporal Supervised Contrastive Learning \\ for Modeling Patient Risk Progression\vspace{-.25em}}

\author{%
\Name{Shahriar Noroozizadeh}
\Email{snoroozi@cs.cmu.edu}\\
\addr Carnegie Mellon University, Pittsburgh, PA, USA
\\
\Name{Jeremy C. Weiss}
\Email{jeremy.weiss@nih.gov}\\
\addr National Library of Medicine, Bethesda, MD, USA
\\
\Name{George H. Chen}
\Email{georgechen@cmu.edu}\\
\addr Carnegie Mellon University, Pittsburgh, PA, USA
\vspace{-3.5em}
}

\begin{document}

\maketitle

\begin{abstract}
We consider the problem of predicting how the likelihood of an outcome of interest for a patient changes over time as we observe more of the patient's data.
To solve this problem, we propose a supervised contrastive learning framework
that learns an embedding representation for each time step of a patient time series.
Our framework learns the embedding space to have the following properties:
(1)~nearby points in the embedding space have similar predicted class probabilities,
(2)~adjacent time steps of the same time series map to nearby points in the embedding space,
and (3)~time steps with very different raw feature vectors map to far apart regions of the embedding space.
To achieve property (3), we employ a \emph{nearest neighbor pairing} mechanism in the raw feature space. 
This mechanism also serves as an alternative to ``data augmentation'', a key ingredient of contrastive learning, which lacks a standard procedure that is adequately realistic for clinical tabular data, to our knowledge.
We demonstrate that our approach outperforms state-of-the-art baselines in predicting mortality of septic patients (MIMIC-III dataset) and tracking progression of cognitive impairment (ADNI dataset). Our method also consistently recovers the correct synthetic dataset embedding structure across experiments, a feat not achieved by baselines. 
Our ablation experiments show the pivotal role of our nearest neighbor pairing.
\end{abstract}

\setlength{\abovedisplayskip}{4pt plus 1.5pt}
\setlength{\belowdisplayskip}{4pt plus 1.5pt}
\setlength{\abovedisplayshortskip}{2pt plus 1.5pt}
\setlength{\belowdisplayshortskip}{2pt plus 1.5pt} 

\vspace{-1em}
\section{Introduction}
\vspace{-.25em}
Modeling disease progression patterns of patients is crucial for developing treatment strategies. Understanding and learning these patterns from longitudinal tabular data, commonly found in healthcare, can be challenging. These time series often vary in length, exhibit irregular sampling, and have many missing entries.
To this end, various methods have emerged in recent years to model such tabular time series data. These methods can broadly be categorized into those that focus only on predicting patient outcomes (e.g., \citealt{choi2016retain, ma2017dipole, devlin2018bert, mollura2021novel}), and those that jointly cluster patients and predict their outcomes (e.g., \citealt{lee2020temporal, lee2020outcome, huang2021deep, carr2021longitudinal, aguiar2022learning, qin2023t}).

The mentioned approaches have several limitations. Specifically, prediction-focused models often struggle to differentiate patients with differing characteristics but the same outcome. 
For instance, predicting in-hospital mortality from patient time series may yield the same risk for an elderly patient with liver dysfunction and a young patient with renal failure, despite them needing different treatments.
In such a case, coarse prediction labels (e.g., in-hospital mortality which encompass diverse underlying causes) do not encourage prediction-centric models to capture clinically relevant details. This is because the model is not required to recognize the semantic differences between these two patients directly to achieve high prediction power.
We term this challenge as capturing ``raw feature heterogeneity'', where distinguishing patients with identical classification outcomes and contrasting raw features becomes a difficulty.

For models that jointly cluster patients and predict their outcomes, these models tend to be less accurate than models solely focused on prediction.
For example, in the supervised temporal clustering models by \citet{lee2020temporal} and \citet{aguiar2022learning}, a prediction neural network is first initialized, after which a clustering module is then added to approximate the encoding representation of the prediction model. 
This approximation incurs some loss in information such that using a patient's assigned cluster to predict the patient's outcome has worse accuracy than using the original prediction network.

The supervised temporal clustering models by \citet{lee2020temporal} and \citet{aguiar2022learning} are also learned in a manner that arguably depends too much on user-specified design choices.
Specifically these models employ k-means for initializing clusters, where the user must specify the number of clusters to use. Using a different clustering algorithm or number of clusters could drastically change the results. This situation may demand costly model re-training if a user finds the learned clusters to be hard to interpret, or either too fine or too coarse.

\vspace{.2em}
\noindent
\textbf{Our contributions.}
To address the issue of raw feature heterogeneity (distinguishing between patients with the same outcome who look quite different) and to avoid the issues encountered with clustering models, our main contribution in this paper is to propose a framework for modeling variable-length tabular time series data called \emph{Temporal Supervised Contrastive Learning} (\mbox{\textsc{Temporal-SCL}}), which does not use clustering during model training due to the challenges stated earlier. 

Our core innovation lies in adopting an embedding-centric paradigm that builds off from the fundamental concept of \emph{contrastive learning} \citep{le2020contrastive} which the goal is to learn an embedding representation where ``similar'' data points exhibit proximate embedding vectors, while ``dissimilar'' data point have distant embeddings. Building on top of \cite{khosla2020supervised}, our embedding-centric approach allows us to impose a structured representation that encapsulates the crucial properties outlined in the introduction for capturing the raw feature heterogeneity while maintaining high predictive power.

\textsc{Temporal-SCL} learns an embedding vector at each time step of a time series. The embedding space has the following properties:
\begin{itemize}[itemsep=0pt,topsep=0pt,parsep=0pt,partopsep=0em,leftmargin=*]
\item[$\bullet$] (Predictive) Each time step's embedding vector helps predict classification outcomes for both the \emph{static} outcome case where a single patient time series is associated with a single class, and the \emph{dynamic} outcome case where a single patient time series has a time-varying classification label.
\item[$\bullet$] (Temporally smooth) Embedding vectors of adjacent time steps within the same patient time series tend to be near each other.
\item[$\bullet$] (Diverse in capturing raw feature heterogeneity) Embedding vectors of dissimilar raw inputs tend to map to separate regions of the embedding space, \emph{even when sharing the same classification outcome}.
\end{itemize}
We achieve the last property using a nearest neighbor pairing mechanism.

We show that \textsc{Temporal-SCL} works well in practice. In two real clinical datasets, \textsc{Temporal-SCL} outperforms various baselines including state-of-the-art methods such as transformers, and removing the nearest neighbor pairing mechanism leads to noticeably lower accuracy. In a synthetic dataset with known ground truth embedding space structure, \mbox{\textsc{Temporal-SCL}} consistently recovers the correct structure (100\% success in 10 experimental repeats with different random seeds) whereas no baseline tested achieves this; removing nearest neighbor pairing makes our method's correct recovery rate~0\%.

We also propose a clustering-based heatmap visualization of the learned embedding space, relating it to raw features and to prediction outcomes. The clustering involved happens only \emph{after} model training and is purely for the purposes of visualization, where our visualization strategy is agnostic to the choice of clustering algorithm used or the number of clusters.

\vspace{-1em}
\section{Background}
\label{sec:background}
\vspace{-.25em}
We state the time series prediction setup we study in Section~\ref{sec:problem-setup}. Our proposed method is based on supervised contrastive learning \citep{khosla2020supervised}, which we review in Section~\ref{sec:supcon}. For any positive integer $k$, we regularly use the notation $[k]\triangleq \{1,2,\dots,k\}$.

\vspace{-1em}
\subsection{Problem Setup}
\label{sec:problem-setup}
\vspace{-.25em}

\noindent
\textbf{Training data.}
We assume that the training data consist of $N$ patients with different time series.
For the \mbox{$i$-th} patient (with $i\in[N]$), we observe $L_i$ time steps, where at each time step, we keep track of $D$ features (e.g., clinical measurements). Specifically, we denote the \mbox{$i$-th} patient's feature vector at time step $\ell\in[L_i]$ (sorted chronologically) as $\mathbf{x}_i^{(\ell)}\in\mathbb{R}^{D}$. 
Moreover, we know the time-step times, where the \mbox{$i$-th} patient's time at the $\ell$-th time step is $t_i^{(\ell)}\in\mathbb{R}$.
This notation allows for features to be static, i.e., a feature could stay constant across time.

We further assume that every time step of a time series belongs to one of $C$ different classes, i.e., the set of classes is $[C]$. In the dynamic outcome case, where the classification label varies over time, we assume that we know the classification label $y_i^{(\ell)}\in[C]$ for every patient
$i\in[N]$ for every time step $\ell\in[L_i]$. For the static outcome case, we only know the classification label at the final time step per time series: we know $y_i^{(L_i)}\in[C]$ for all $i\in[N]$, and at earlier time steps $\ell < L_i$, we set $y_i^{(\ell)}=``?"$, which could be thought of as an additional ``unknown'' class.

\vspace{.2em}
\noindent
\textbf{Prediction task.} Given a test time series, suppose that we observe the feature vector $\mathbf{x}_*\in\mathbb{R}^{D}$ at a \emph{single time step}. We aim to predict the target label $y_*\in[C]$ corresponding to $\mathbf{x}_*$. In the dynamic outcome setting, this target label is for the same time step as $\mathbf{x}_*$. In the static outcome setting, this target label is for the label of final time step in the test time series.

We focus on predicting the classification outcome for each time step.
This captures the extreme case of a test time series consisting of a single time step. 
For example, this could happen in a clinical setup if a patient enters a new hospital for which we see measurements of this patient for the first time.
Our real-world clinical dataset experiments later reveal that our model outperforms baselines that leverage complete historical observations.

Importantly, although we focus on prediction given a single time step's feature vector $\mathbf{x}_*$, our problem setup allows for a feature in $\mathbf{x}_*$ to capture historical information (e.g., a feature in $\mathbf{x}_*$ could be the maximum observed white blood count over the past 4 hours).

\vspace{-1em}
\subsection{Supervised Contrastive Learning (SCL)}
\label{sec:supcon}
\vspace{-.25em}

SCL learns an embedding representation of the data so that ``similar'' data points have embedding vectors that have higher cosine similarity compared to those of ``dissimilar'' data points. Data points are ``similar'' if they have the same classification label. As this framework was originally developed without temporal structure, we drop superscripts previously used to indicate dependence on time.

\vspace{.2em}
\noindent
\textbf{Notation.}
We use $f$ to denote the so-called \emph{encoder network}, where given any point~$\mathbf{x}$ in the raw input feature space, its embedding representation is $f(\mathbf{x})$. This embedding representation is constrained to be a $d$-dimensional Euclidean vector with norm~1, also referred to as a \emph{hyperspherical embedding}. Thus, $\|f(\mathbf{x})\|=1$. We denote this hyperspherical output space as $\mathcal{S}^{d-1}\triangleq \{\mathbf{z}\in\mathbb{R}^d\text{ s.t.~}\|\mathbf{z}\|=1\}$. For $\mathbf{u},\mathbf{v}\in\mathcal{S}^{d-1}$, the cosine similarity between $\mathbf{u}$ and $\mathbf{v}$ is $\frac{\langle \mathbf{u}, \mathbf{v} \rangle}{\|\mathbf{u}\|\|\mathbf{v}\|} = \langle \mathbf{u}, \mathbf{v} \rangle$, where $\langle\cdot,\cdot\rangle$ denotes the dot product.

We also denote $g$ as the \emph{predictor network} that maps the embedding output of the encoder in $\mathcal{S}^{d-1}$ to a probability distribution space over $C$ classes.

\vspace{.4em}
\noindent
\textbf{\bfseries\scshape Simple-SCL.}
For ease of exposition, we present a simplified version of SCL that we call \textsc{Simple-SCL}, \emph{which does not use data augmentation} (since our experiments later will be on tabular data without data augmentation as generating realistic fake patient data is challenging in the clinical setting).
We learn the encoder $f$ using minibatch gradient descent. For a minibatch of $B$ training points $\mathbf{x}_1,\dots,\mathbf{x}_B$ with corresponding classification labels $y_1,\dots,y_B$, we denote the embedding vectors of these points as $\mathbf{z}_1=f(\mathbf{x}_1),\dots,\mathbf{z}_B=f(\mathbf{x}_B)$. Since we want points with the same label to have high cosine similarity, we keep track of which points have the same label. To do this, we let $\mathcal{P}(i)$ denote the set of points with the same label as the $i$-th point, excluding the $i$-th point:
\begin{equation}
\mathcal{P}(i) \triangleq  \{ j\in[B]\text{ s.t.~}y_j=y_i\text{ and }j\ne i \}.
\label{eq:neighbors-of-i}
\end{equation}
Next, we define the following ratio:
\begin{equation}
\Psi(i,j;\tau)
\triangleq  \frac{\exp(\langle \mathbf{z}_i, \mathbf{z}_j \rangle / \tau)}{\sum_{k\in[B]\text{ s.t.~}k\ne i}\exp(\langle \mathbf{z}_i, \mathbf{z}_k \rangle / \tau)},
\label{eq:Psi}
\end{equation}
where the constant $\tau>0$ is a user-specified hyperparameter. 
The key idea is that if the $i$-th and $j$-th points have the same label (so $j\in\mathcal{P}(i)$), then we want $\Psi(i,j;\tau)$ to be large: the numerator being large means that cosine similarity $\langle\mathbf{z}_i, \mathbf{z}_j\rangle$ is large while the denominator provides a normalization to ensure that $\Psi(i,j;\tau)\in[0,1]$. Then to encourage $\Psi(i,j;\tau)$ to be large for all $i\in[B]$ and $j\in\mathcal{P}(i)$, we minimize the~loss
\begin{equation}
L_{\text{Simple-SCL}}
\triangleq 
-\!\!\!\!\!\!\!\!\!\!\!\!\!\!\!\sum_{i\in[B]\text{ s.t.~}|\mathcal{P}(i)|\ge1}
   \frac{1}{|\mathcal{P}(i)|}
     \sum_{j\in\mathcal{P}(i)} \log\Psi(i,j;\tau).
\label{eq:sup-con-loss}
\end{equation}
The original SCL uses a loss of the same form but with data augmentation (see Appendix~\ref{sec:relating-to-original-SCL}).

Lastly, in line with \cite{khosla2020supervised}, after the encoder is learned, we train a predictor network $g$ to map the resulting embeddings to a probability distribution over all classes.

\vspace{-1em}
\section{Method}
\vspace{-.25em}
We now introduce \textsc{Temporal-SCL}, our temporal adaptation of SCL. It consists of three networks: an encoder $f$, a predictor $g$, and a temporal network $h$. 
We give a high-level overview of these networks in Section~\ref{sec:overview-networks} and how they are trained and used for prediction in Section~\ref{sec:training}. We propose a method for visualizing embedding vectors in Section~\ref{sec:visualizing}.

\vspace{-1em}
\subsection{Overview of \textsc{Temporal-SCL}'s Networks}
\label{sec:overview-networks}
\vspace{-.25em}

\begin{figure}[t!]
\floatconts
    {fig:model-overview}
    {\caption{Overview of \textsc{Temporal-SCL}}\vspace{-1em}}
    {%
        \vspace{.4em}
        \subfigure[Encode patients at time step level]{%
            \label{fig:encoding}
            \centering
            \includegraphics[width=0.8\linewidth,trim={6pt 0pt 6pt 6pt},clip]{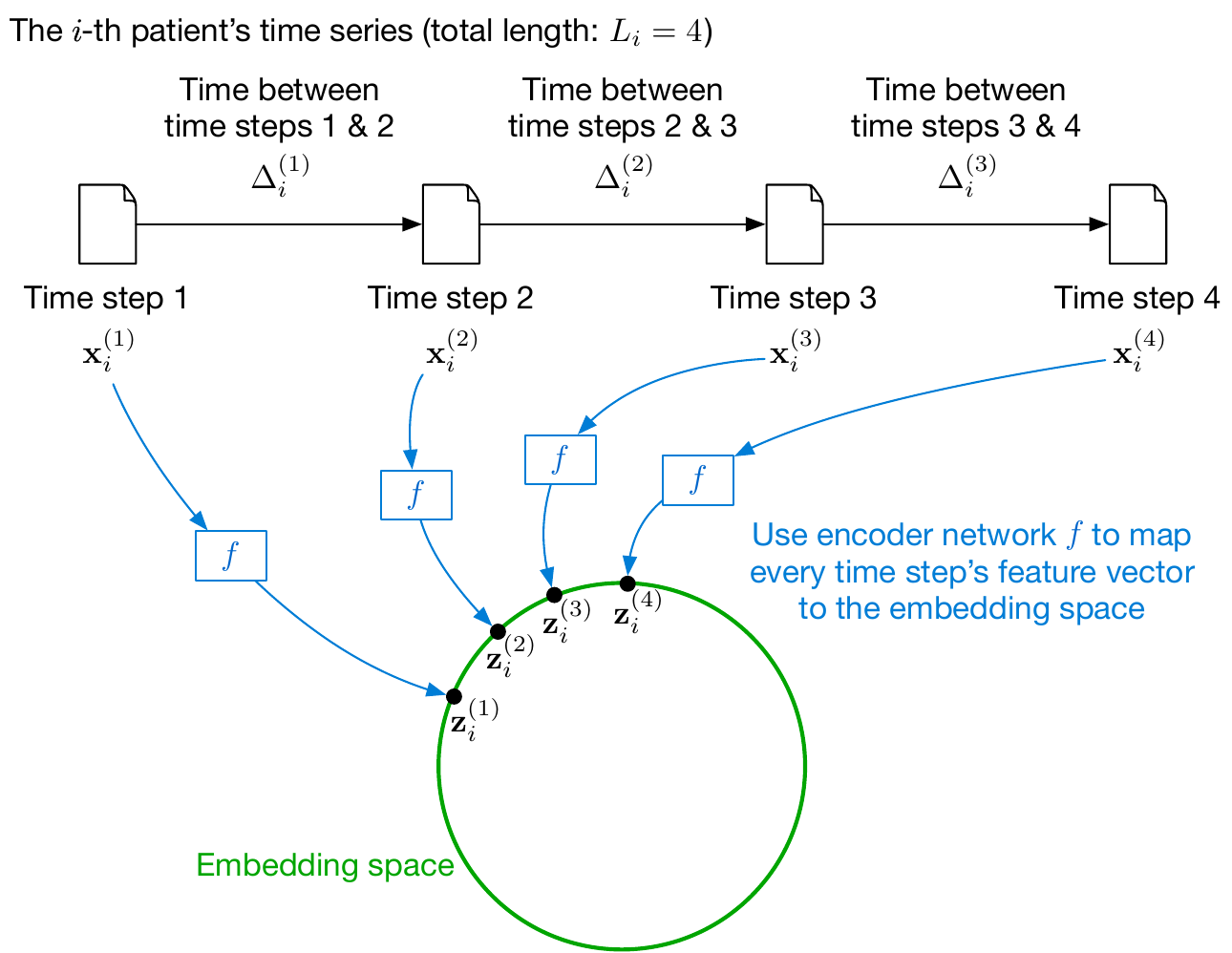}
        } \\[.5em]
        \subfigure[Encourage temporal smoothness]{%
            \label{fig:temporal-smoothing}
            \centering
            \includegraphics[width=0.8\linewidth,trim={6pt 0pt 6pt 6pt},clip]{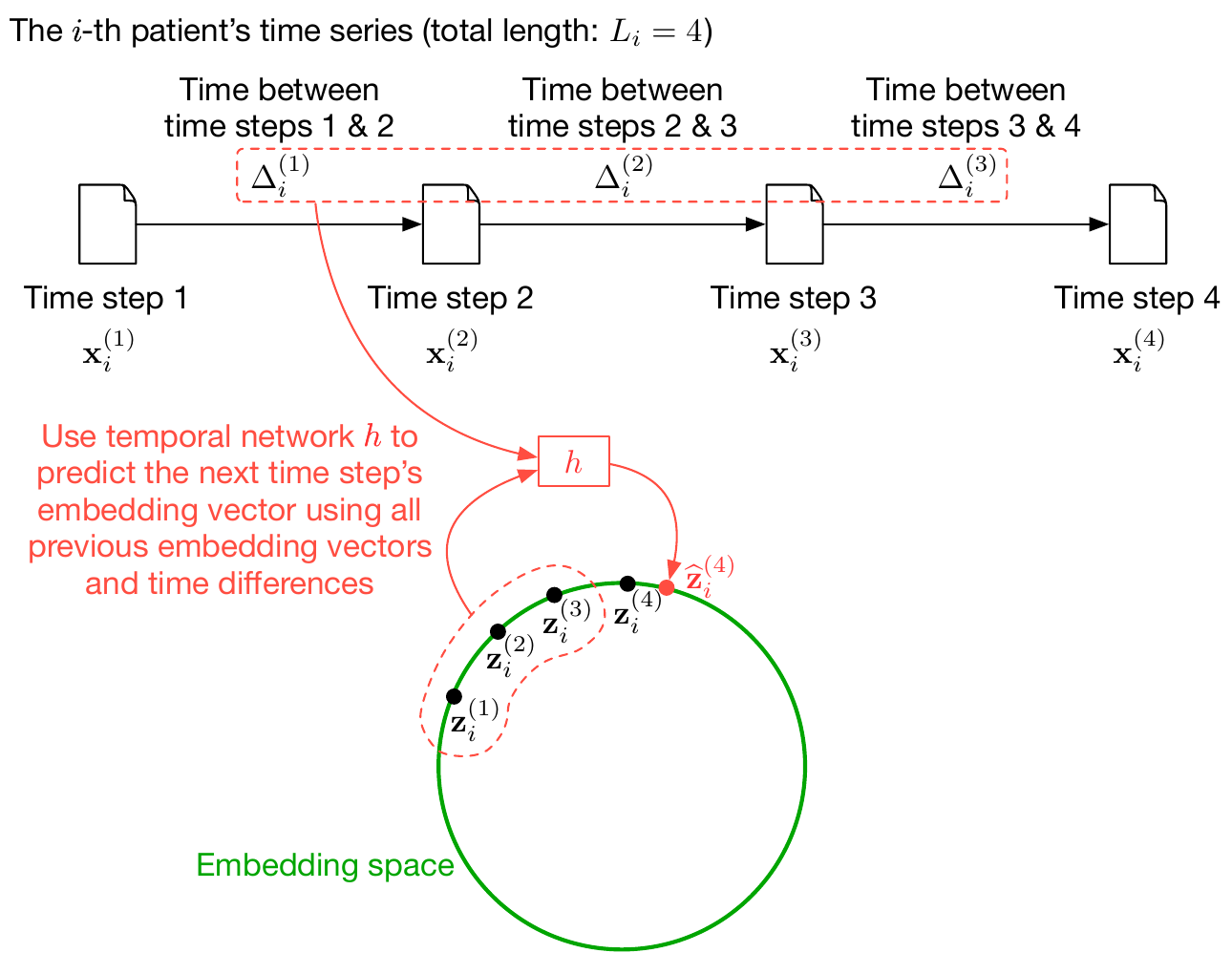}
        } \\[0em]
        \subfigure[Predict using encoded representation]{%
            \label{fig:prediction}
            \centering
            \includegraphics[width=0.8\linewidth,trim={6pt 0pt 6pt 21em},clip]{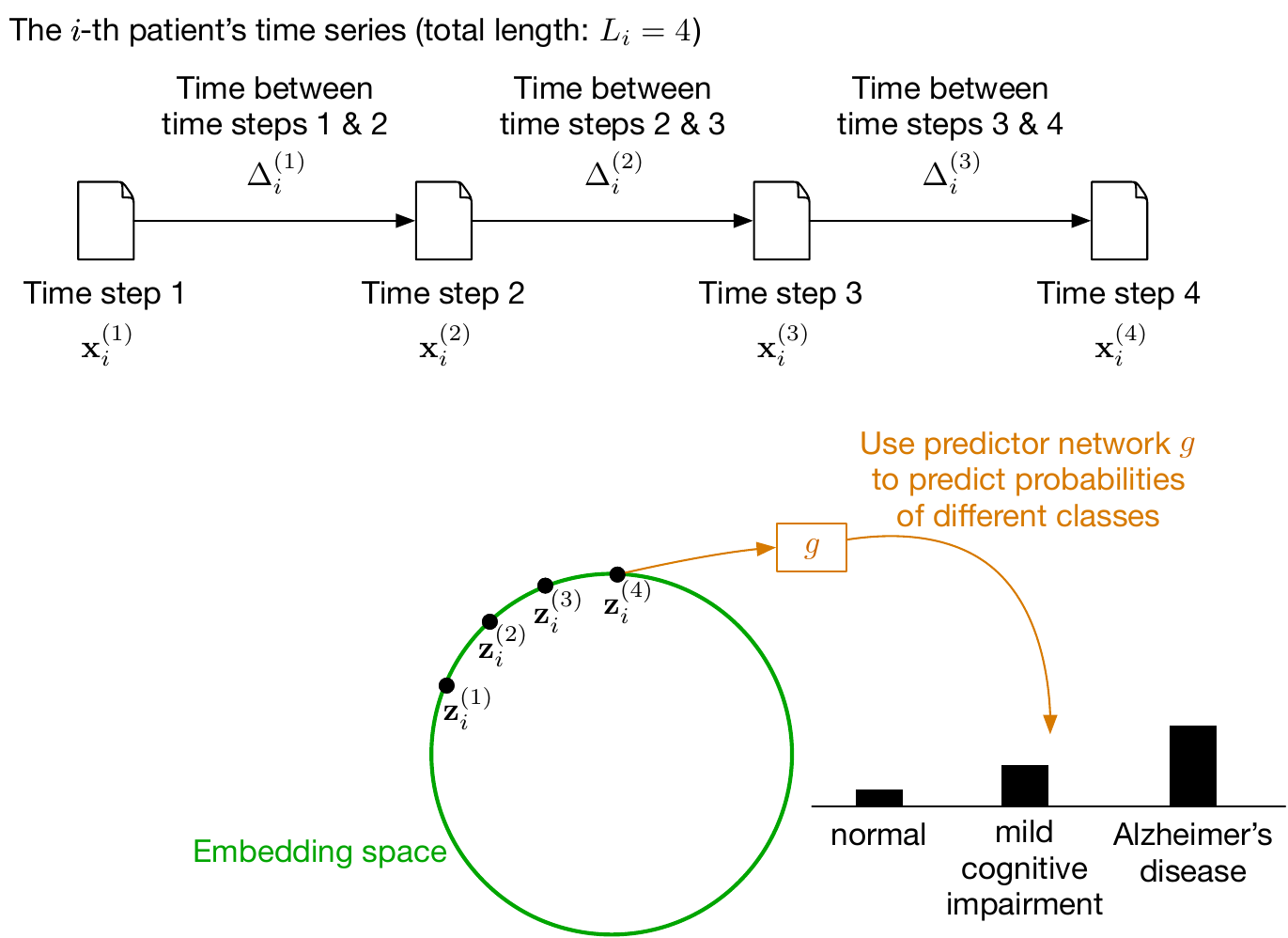}
        }
        \vspace{-1.75em}
    }
\end{figure}

\noindent
\textbf{Encoder network $f$.}
We aim to learn an embedding representation of every time step of each time series. 
Just as in Section~\ref{sec:supcon}, this amounts to learning an encoder network $f$ that maps from the raw feature vector space (now for just a single time step) to the hyperspherical space~$\mathcal{S}^{d-1}$ (Fig.~\ref{fig:encoding}).
Since we now account for time steps, we add superscripts: we let $\mathbf{z}_i^{(\ell)}\triangleq f(\mathbf{x}_i^{(\ell)})$ denote the embedding vector of the \mbox{$i$-th}  data point's feature vector at time step~$\ell$. 

Aside from using time steps, the major difference between \textsc{Temporal-SCL} and \textsc{Simple-SCL} is that for \textsc{Temporal-SCL}, feature vectors are considered similar if they simultaneously have similar outcomes (i.e., classification labels $y_i^{(\ell)}$) \emph{and} similar input feature vectors (unlike with \textsc{Simple-SCL}, where similarity is purely based on classification labels).

\vspace{.2em}
\noindent
\textbf{Predictor network $g$.}
To ensure the embedding space's predictive power of the outcome, we train predictor network $g$ mapping $\mathcal{S}^{d-1}$ to a probability distribution space over $C$ classes (Fig.~\ref{fig:prediction}).

\vspace{.2em}
\noindent
\textbf{Temporal network $h$.}
To encourage two adjacent time steps in the same time series $\mathbf{x}_i^{(\ell)}$ and $\mathbf{x}_i^{(\ell+1)}$ to map to embedding vectors $\mathbf{z}_i^{(\ell)}$ and $\mathbf{z}_i^{(\ell+1)}$ that are closeby, we learn a temporal network $h$ that predicts how embedding vectors change over time (Fig.~\ref{fig:temporal-smoothing}). 
For the \mbox{$i$-th} training time series, we define the duration $\delta_i^{(\ell)}\triangleq t_i^{(\ell+1)} - t_i^{(\ell)}$ for $\ell\in[L_i-1]$. Then $h$ takes as input the sequence $(\mathbf{z}_i^{(1)},\delta_i^{(1)}),\dots,(\mathbf{z}_i^{(\ell)},\delta_i^{(\ell)})$ and outputs a prediction for $\mathbf{z}_i^{(\ell+1)}$.
We ask that $\mathbf{z}_i^{(\ell+1)}$ and $h\big((\mathbf{z}_i^{(1)},\delta_i^{(1)}),\dots,(\mathbf{z}_i^{(\ell)},\delta_i^{(\ell)})\big)$ be close (using squared Euclidean distance loss). Thus, $h$ aims to make the next time step's embedding vector predictable based on all previous time steps and could be thought of as a regularization term. 
A similar temporal regularization strategy was used by~\citet{lee2019dynamic}.

\vspace{-1em}
\subsection{Overview of Training and Prediction}
\label{sec:training}
\vspace{-.25em} 

\noindent
\textbf{Training.}
We train \textsc{Temporal-SCL} in three phases:
\begin{enumerate}[leftmargin=*,itemsep=0pt,parsep=0pt,topsep=2pt]

\item (Pre-training) We initialize the embedding space by pre-training the encoder~$f$ using data at the individual time step level and \mbox{\textsc{Simple-SCL}}. 
Note that during this phase, we do not model nor use temporal structure, and we effectively treat the different time steps as separate. We explain this phase in more detail in Section~\ref{sec:learning-the-encoder}, and our experiments later show that this phase significantly improves the model's prediction accuracy.

\item (Encoder and temporal network training) After pre-training the encoder $f$, we account for temporal structure by jointly training $f$ and the temporal network $h$. Details of this phase are in Section~\ref{sec:transform}.

\item (Learning the predictor network) At this point, we treat the encoder $f$ as fixed, so we can compute all the training embedding vectors at different time steps (the $\mathbf{z}_i^{(\ell)}$ variables). In the dynamic outcome case, we learn the predictor network $g$ by treating the $\mathbf{z}_i^{(\ell)}$ variables as input feature vectors and the corresponding $y_i^{(\ell)}$ variables as target labels, minimizing cross-entropy loss. In the static outcome case, we instead set the target label for $\mathbf{z}_i^{(\ell)}$ to be the final time step's label $y_i^{(L_i)}$. As this phase amounts to standard neural net classifier training with cross-entropy loss, we omit further details.

\end{enumerate}
For ease of exposition, we explain the first two phases in Sections~\ref{sec:learning-the-encoder} and~\ref{sec:transform} only for the dynamic outcome case that we described in Section~\ref{sec:problem-setup}; the static outcome case is very similar and covered in Appendix~\ref{sec:handling-static-outcome}.

\vspace{.2em}
\noindent
\textbf{Prediction.}
At test time, we are given a feature vector $\mathbf{x}_*\in\mathbb{R}^{D}$ for a single time step. To make a prediction for~$\mathbf{x}_*$, we first compute the embedding vector of $\mathbf{x}_*$ given by $\mathbf{z}_* \triangleq  f(\mathbf{x}_*)$. Then the predicted class probabilities are precisely given by $g(\mathbf{z}_*)$.

\vspace{-1em}
\subsection{Pre-training the Encoder}
\label{sec:learning-the-encoder}
\vspace{-.25em}

We now explain how we adapt \textsc{Simple-SCL} to time series data and to encourage learned embedding vectors to be diverse in capturing raw feature heterogeneity. The latter uses a \emph{nearest neighbor pairing} mechanism.
We first pre-train our encoder network $f$ following a similar process as the \textsc{Simple-SCL} encoder, with specific modifications described below.

\vspace{.2em}
\noindent
\textbf{\textsc{Simple-SCL} encoder with time steps.}
Encoder of \textsc{Temporal-SCL}, $f$, has the same structure as \textsc{Simple-SCL} encoder. We define every time step of every time series as its own \emph{snapshot}:  $(\mathbf{x}_i^{(\ell)}, y_i^{(\ell)})$.
Thus, we take the ``training data'' of our encoder network to be all the snapshots: $\bigcup_{i=1}^N \bigcup_{\ell=1}^{L_i} \{ (\mathbf{x}_i^{(\ell)}, y_i^{(\ell)}) \}$.

\vspace{.2em}
\noindent
\textbf{Nearest neighbor pairing for identifying similar snapshots.}
\label{sec:nearest-neighbor-refinement}
Whereas \textsc{Simple-SCL} (that did not have time steps) considered two points to be similar if they share the same classification outcome, we now instead consider two \emph{snapshots} to be similar if they share the same classification outcome \emph{and} their raw feature vectors are ``close to each other''. We use random sampling to find pairs of similar snapshots:
\begin{enumerate}[leftmargin=*,itemsep=0pt,topsep=1pt,parsep=0pt]
\small
\item Initialize the set of snapshot pairs to be empty: $\mathcal{E} \leftarrow \emptyset$.
\item For each class $c\in[C]$:
\begin{enumerate}[leftmargin=8pt,itemsep=0pt,topsep=0pt,parsep=0pt]
\item 
Let the set $\mathcal{A}_c$ consist of all snapshots with label $c$.
\item While $|\mathcal{A}_c| \ge 2$:
\begin{enumerate}[leftmargin=10pt,itemsep=0pt,topsep=0pt,parsep=0pt]
\item Choose snapshot $(\mathbf{x}_i^{(\ell)}, y_i^{(\ell)})$ randomly from $\mathcal{A}_c$.
\item (Nearest neighbor search) Among the other snapshots in $\mathcal{A}_c$, find the one whose feature vector is closest to $\mathbf{x}_i^{(\ell)}$ (e.g., using Euclidean distance). Denote the snapshot found as~$(\mathbf{x}_{i'}^{(\ell')}, y_{i'}^{(\ell')})$.
\item Add snapshot pair $\big((\mathbf{x}_i^{(\ell)}, y_i^{(\ell)}),(\mathbf{x}_{i'}^{(\ell')}, y_{i'}^{(\ell')})\big)$ to~$\mathcal{E}$.
\item Remove $(\mathbf{x}_i^{(\ell)}, y_i^{(\ell)})$ and $(\mathbf{x}_{i'}^{(\ell')}, y_{i'}^{(\ell')})$ from $\mathcal{A}_c$.
\end{enumerate}
\end{enumerate}
\end{enumerate}
To sample a minibatch of $B$ data points for minibatch gradient descent, where we assume that $B$ is even, we randomly choose $B/2$ pairs from the set $\mathcal{E}$; denote the set of these $B/2$ pairs as $\mathcal{E}_{\text{batch}}$. Note that the $B/2$ pairs in $\mathcal{E}_{\text{batch}}$ correspond to a total of $B$ different snapshots; denote the set of these $B$ snapshots as $\mathcal{V}_{\text{batch}}$. Then the loss we use for minibatch gradient descent during pre-training is
\begin{align*}
    &L_{\text{SCL-snapshots}} \triangleq \\
    &- \!\!\!\!\!\!\!\!
    \sum_{\substack{((\mathbf{x}_i^{(\ell)}, y_i^{(\ell)}), \\ ~\!(\mathbf{x}_{i'}^{(\ell')}, y_{i'}^{(\ell')})) \\ ~\!\in~\!  \mathcal{E}_{\text{batch}}}} \!\!\!\!\!\!
    \log\Bigg[
\frac{\exp(\langle f(\mathbf{x}_i^{(\ell)}), f(\mathbf{x}_{i'}^{(\ell')}) \rangle / \tau)}{\sum\limits_{\substack{(\mathbf{x}_{i''}^{(\ell'')}, y_{i''}^{(\ell'')}) ~\!\in~\!  \mathcal{V}_{\text{batch}} \\ \text{~s.t.~}(i'',\ell'')\ne(i,\ell)}} \!\!\!\!\!\!\!\!\!\!\!\!\!\!\!\exp(\langle f(\mathbf{x}_i^{(\ell)}), f(\mathbf{x}_{i''}^{(\ell'')}) \rangle / \tau)}\Bigg]
\end{align*}
For simplicity, our experiments later use Euclidean distance to find nearest neighbors (we standardize features to control for scale differences), efficiently computed using fast approximate nearest neighbor search software \citep{malkov2018efficient}.

\begin{figure*}[t!]
\vspace{-.25em}
\includegraphics[width=1\linewidth]{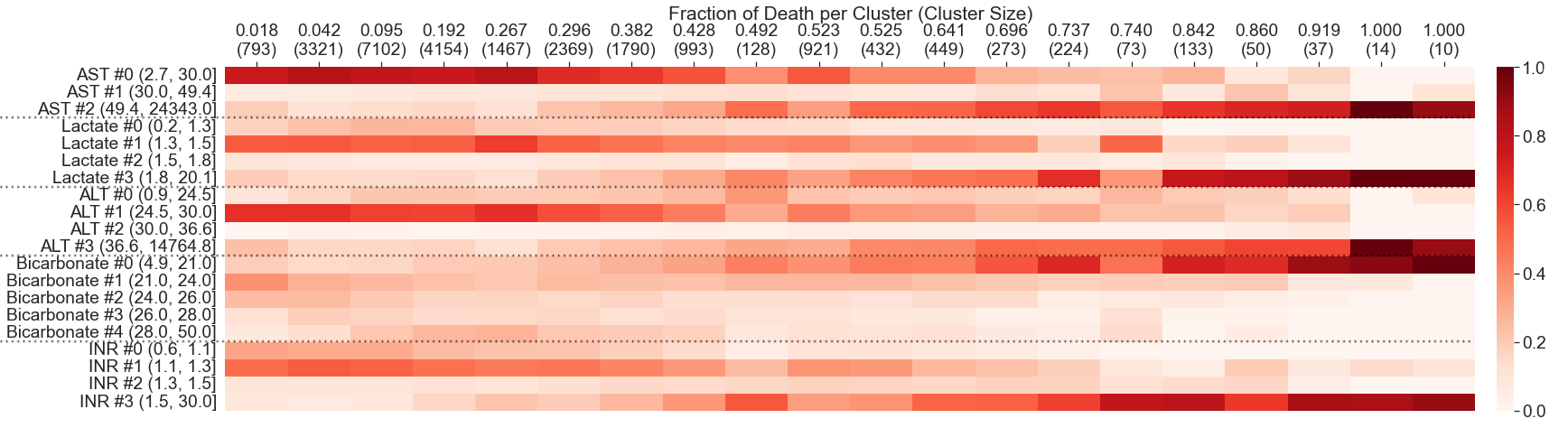}
\vspace{-2.5em}
 \caption{Heatmap showing how features (rows) vary across clusters (columns) for the sepsis cohort of the MIMIC dataset. Heatmap intensity values can be thought of as the conditional probability of seeing a feature value (row) conditioned on being in a cluster (column); these probabilities are estimated using test set snapshots. Columns are ordered left to right in increasing fraction of test set snapshots that come from a time series that has a final outcome of death.\vspace{-2em}
 }   \label{fig:heatmapMIMIC}
\end{figure*}

\noindent
\textbf{Ablation.}
Later on in our experiments, we conduct ablation experiments where we do not use nearest neighbor pairing. The only change is that in step 2(b), steps i.~and ii.~are replaced by randomly choosing two different snapshots $(\mathbf{x}_{i}^{(\ell)}, y_{i}^{(\ell)})$ and $(\mathbf{x}_{i'}^{(\ell')}, y_{i'}^{(\ell')})$ from $\mathcal{A}_c$ to pair up (uniformly at random).

\vspace{.2em}
\noindent
\textbf{Note on data augmentation.}
A key component of contrastive learning is \emph{data augmentation} (see the review by \citet{le2020contrastive}), a procedure that creates perturbed versions of data points (e.g., when working with images, we could randomly rotate, translate, and crop an image). Since we work with tabular data, a challenge arises in that to the best of our knowledge, there is no widely accepted, clinically suitable data augmentation method for such data. Thus, our presentation of \textsc{Temporal-SCL} is \emph{without data augmentation}; in fact, the nearest neighbor pairing mechanism proposed above is also a substitute for data augmentation. 
While we do not use data augmentation, we point out that if it is available, adding it is trivial. The simplest way is that before each training epoch, we apply data augmentation to the original training dataset to get a fresh augmented training dataset, which we use with our nearest neighbor pairing procedure. More details are in Appendix~\ref{sec:data-augmentation-with-nn-pairing}.

\vspace{-1em}
\subsection{Encoder and Temporal Network Training}
\label{sec:joint-optimization}
\vspace{-.5em}
 
\label{sec:transform}

In the second stage of training a \textsc{Temporal-SCL} model, we jointly train the encoder and temporal networks by minimizing the overall loss
\begin{align}
    L_{\text{overall}} &=L_{\text{SCL-snapshots}}+ \alpha L_{\text{temp-reg}},
\label{eq:mainLoss}
\end{align}
where $L_{\text{temp-reg}}$ is a temporal smoothness loss term (to be defined shortly), and ${\alpha\ge0}$ is a hyperparameter that trades off between the two losses on the right-hand side.
Recall that for the \mbox{$i$-th} training data point, we previously defined the time duration $\delta_i^{(\ell)}\triangleq t_i^{(\ell+1)} - t_i^{(\ell)}$ for $\ell=1,2,\dots,{L_i-1}$. 
Moreover, $h$ takes as input the sequence $\{(\mathbf{z}_i^{(t)},\delta_i^{(t)})\}_{t=1}^{\ell}$ and outputs a prediction for $\mathbf{z}_i^{(\ell+1)}$. 
We ask that $\mathbf{z}_i^{(\ell+1)}$ and $h\big(\{(\mathbf{z}_i^{(t)},\delta_i^{(t)})\}_{t=1}^{\ell}\big)$ be close by squared Euclidean distance, across all data points and time steps:
\begin{align*}
&L_{\text{temp-reg}}
\!\triangleq \!\!\!\!\!
&\!\frac{1}{N}\!\sum_{i=1}^N  \! \sum_{\ell=1}^{L_i - 1} \! \frac{
\big\| h\big(\{(\mathbf{z}_i^{(t)},\delta_i^{(t)})\}_{t=1}^{\ell}\big)
\!-\! \mathbf{z}_i^{(\ell+1)} \big\|^2 }{L_i-1}.
\end{align*}

\vspace{-1em}
\subsection{Visualizing Embedding Vectors}
\label{sec:visualizing}
\vspace{-.3em}

To find common patterns in the learned embedding space, we cluster on the different snapshots' embedding vectors (the $\mathbf{z}_i^{(\ell)}$ variables) \emph{after} training a \textsc{Temporal-SCL} model. We have found standard agglomerative clustering \citep{murtagh2012algorithms} to work well here and this approach trivially allows the user to adjust the granularity of clusters as needed, even from fitting the clustering model once. The idea is that we start with every snapshot's embedding vector as its own cluster and keep merging the closest two clusters (e.g., using complete linkage to decide on which two clusters are closest) until we are left with a single cluster that contains all the snapshots' embedding vectors. In conducting this procedure, we store each intermediate clustering result. Then the user could use our visualization strategy (to be described next) with any intermediate clustering result, ranging from fine- to coarse-grain clusters.

To visualize \emph{any} cluster assignment of the snapshot embedding vectors (the clustering method need not be agglomerative clustering), we make a heatmap inspired by \citet{li2020neural}, where columns correspond to different clusters and rows correspond to features.\footnote{Note that we discretize continuous features into bins. This discretization is only used during visualization and not used when training the \mbox{\textsc{Temporal-SCL}} model.} The intensity value at the $i$-th row and $j$-th column is the fraction of test set snapshots that have the $i$-th row's feature value among test set snapshots in the $j$-th column's cluster (i.e., a conditional probability of seeing a feature value given being in a cluster). An example is shown in Fig.~\ref{fig:heatmapMIMIC}; details on the dataset used and model training are in Section~\ref{sec:ClinicalExperiments}. 
We show the ``top'' 5 features,
where we rank features based on the maximum observed difference across clusters.\footnote{Per row in the heatmap, compute the difference between the largest and smallest intensity values across the row, and rank rows using these differences, where we keep rows that correspond to the same underlying feature together.}

\vspace{-1.15em}
\section{Experiments}
\vspace{-.35em}
We benchmark \textsc{Temporal-SCL} on tabular time series data:
a synthetic dataset with known ground truth embedding space structure (Section~\ref{sec:ToyExperiment}), and two standard real clinical datasets with unknown embedding space structure (Section~\ref{sec:ClinicalExperiments}). 
We also examine how \textsc{Temporal-SCL} works without pre-training and, separately, without nearest neighbor pairing.

As baselines, we use (a) purely supervised methods: logistic regression, an LSTM~\citep{hochreiter1997long}, \mbox{RETAIN} \citep{choi2016retain}, Dipole \citep{ma2017dipole}, a BERT-based transformer \citep{devlin2018bert}; (b) temporal predictive clustering methods: AC-TPC \citep{lee2020temporal} and T-Phenotype \citep{qin2023t}; and (c) Self-Supervised Learning and Contrastive Learning based model: SMD-SSL \citep{raghu2023sequential} with SimCLR loss \citep{chen2020simple} and VICReg loss \citep{bardes2021vicreg}, and \mbox{\textsc{Simple-SCL}} (treating snapshots as separate data points and without nearest neighbor pairing). Details on baselines are in Appendix~\ref{sec:baselines}.

For all datasets, we randomly split the data into 60\% training, 20\% validation, and 20\% test sets. 
This split is done at the patient level so the presented test set results are all on patients not encountered during training. Details of our training including network types and hyperparameter grids, and an explanation of evaluation metrics are in Appendix~\ref{sec:ExperimentalSetup}.
\footnote{Our code is available at: \\ \url{https://github.com/Shahriarnz14/Temporal-Supervised-Contrastive-Learning}}

\vspace{-1em}   
\subsection{Synthetic Data}
\vspace{-.25em}

\begin{figure*}[!t]
\floatconts
    {fig:toyDataset}
    {\vspace{-2em}\caption{Synthetic dataset: panel (a) shows the only 4 possible time series trajectories (each true embedding vector state has a unique color-shape combination; there are 10 such states); every time series has 3 time steps and belongs to one of two classes \textcolor{red}{red}/\textcolor{blue}{blue}. Panels (b)-(e) show learned embedding spaces of four methods; only \textsc{Temporal-SCL} correctly recovers the 10 ground truth states. A version of this figure with embeddings of all methods evaluated is in Fig.~\ref{fig:toyDataset_result_complete}.\vspace{-1.25em}
    }}
    {%
     \vspace{-.25em}
        \subfigure[Ground truth]{%
            \label{fig:toyDataset_Raw}
            \includegraphics[width=0.195\linewidth]{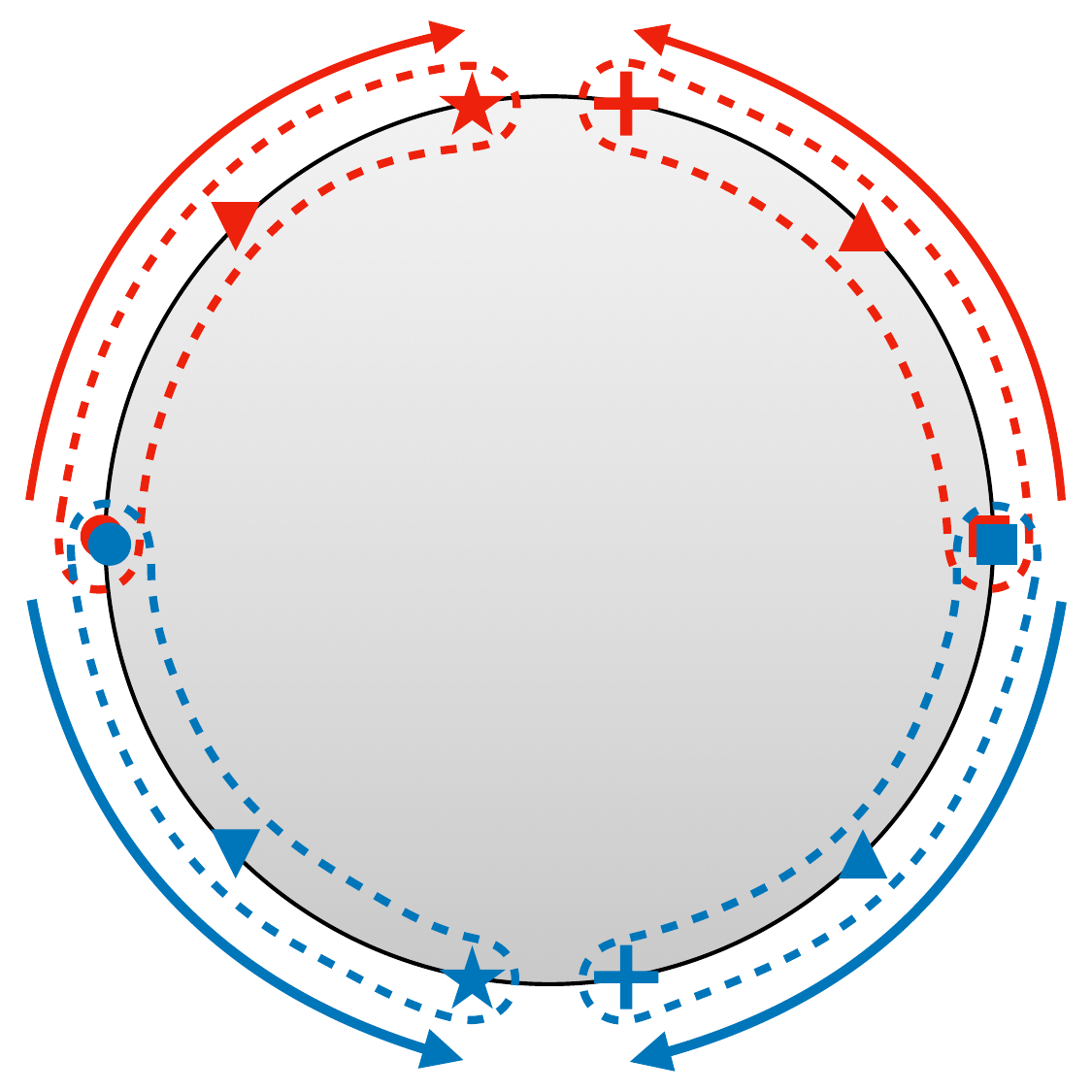}
        }
        \subfigure[Transformer]{%
            \label{fig:synth-transformer}
            \includegraphics[width=0.18\linewidth]{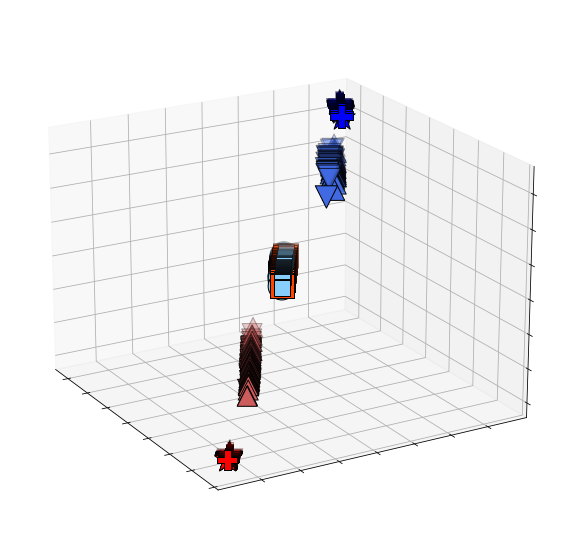}
        }
        \subfigure[T-Phenotype]{%
            \label{fig:synth-tphenotype}
            \includegraphics[width=0.18\linewidth]{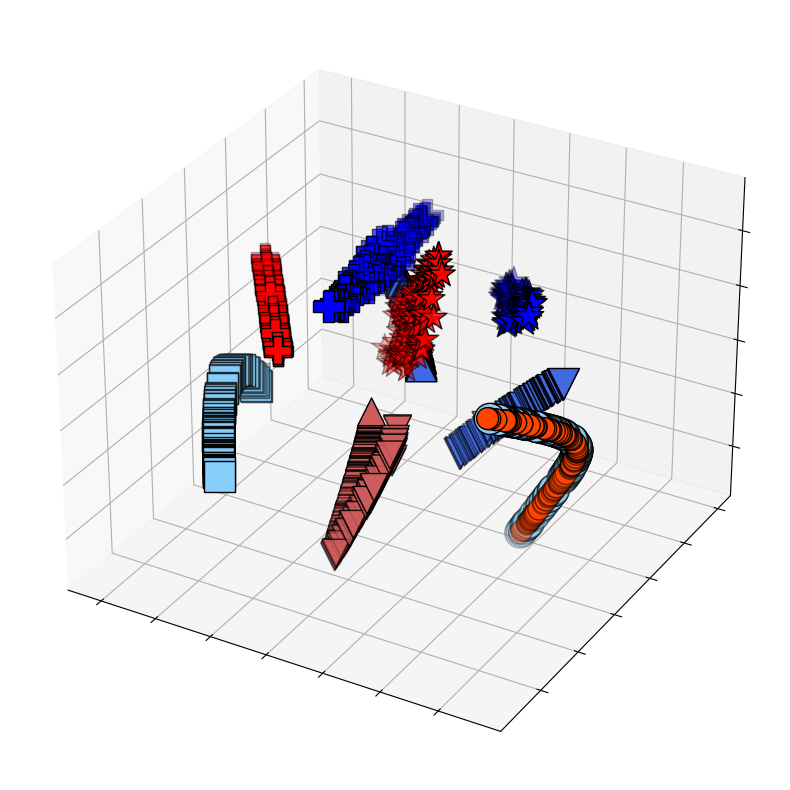}
        }
        \subfigure[\small\textsc{Temporal-SCL} \newline (no NN pairing)]{%
            \label{fig:synth-tscl-no-nnp}
            \includegraphics[width=0.18\linewidth]{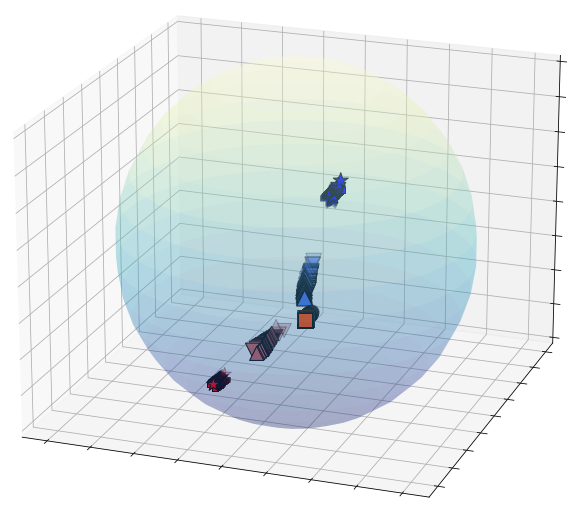}
        }
        ~
        \subfigure[\small\textsc{Temporal-SCL}]{%
            \label{fig:synth-tscl}
            \includegraphics[width=0.18\linewidth]{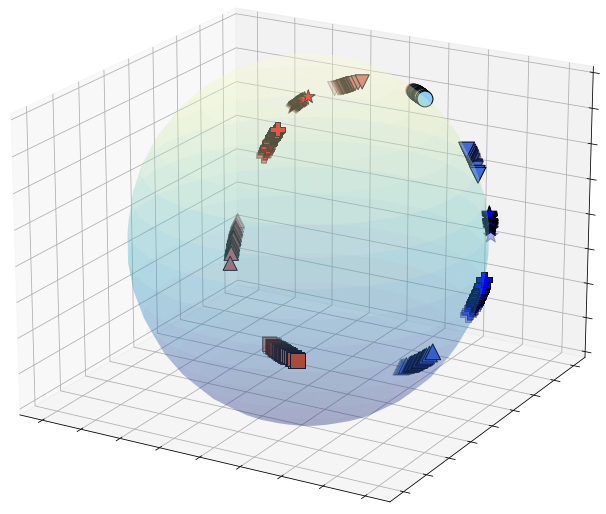}
        }
    }
\end{figure*}
 
\label{sec:ToyExperiment}

\noindent
\textbf{Data.}
We generate a 2D dataset where every patient time series has exactly 3 time steps. For simplicity, we only consider the static outcome case so that each time series has a single label (one of two classes: \textcolor{red}{red} or \textcolor{blue}{blue}). The points are all on a 2D circle, where the only 4 possible time series in the embedding space are shown in Fig.~\ref{fig:toyDataset_Raw}. 
For example, one possible patient time series is ``\scalerel*{\includegraphics{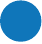}}{B} $\rightarrow$ \scalerel*{\includegraphics{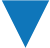}}{B} $\rightarrow$ \scalerel*{\includegraphics{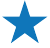}}{B}''. There are a total of 10 true embedding vector locations, which could be thought of as cluster centers. \emph{Note that we take the embedding space and raw feature space to be the same.} When we generate synthetic time series, each point is based on one of the 10 true ground truth embedding vector locations with Gaussian noise added. We randomly sample 200 of each of the 4 possible trajectories so that we have a total of 800 time series. See Appendix~\ref{sec:syntheticData_full} for details.

\begin{table*}[t]
\vspace{-1em}
\caption{Synthetic data test set metrics 
(mean $\pm$ std.~dev.~across 10 experiments) and qualitative assessment of successful ground truth cluster recovery across experiments.}\vspace{-.75em}
\centering
\adjustbox{max width=.85\linewidth}{%
\begin{tabular}{cccccc}
\toprule
Model & AUROC & AUPRC & Silhouette Index & \begin{tabular}[c]{@{}c@{}}Cluster Prediction\\ Accuracy\end{tabular} & Recovery \\ \midrule
Logistic Regression                         & 0.90$\pm$0.01 & 0.90$\pm$0.00 & 0.09$\pm$0.01 & 0.60$\pm$0.06 & \xmark \\
LSTM                                        & 0.95$\pm$0.01 & 0.95$\pm$0.00 & 0.09$\pm$0.01 & 0.60$\pm$0.05 & \xmark \\
RETAIN                                      & 0.95$\pm$0.01 & 0.95$\pm$0.00 & 0.33$\pm$0.06 & 0.69$\pm$0.06 & \omark \\
DIPOLE                                      & 0.95$\pm$0.01 & 0.95$\pm$0.00 & 0.30$\pm$0.06 & 0.69$\pm$0.04 & \omark \\
Transformer:BERT                            & 0.95$\pm$0.01 & 0.95$\pm$0.00 & 0.11$\pm$0.01 & 0.67$\pm$0.03 & \xmark \\
AC-TPC                                      & 0.95$\pm$0.01 & 0.95$\pm$0.00 & 0.09$\pm$0.01 & 0.77$\pm$0.03 & \xmark  \\
T-Phenotype                                 & 0.95$\pm$0.01 & 0.95$\pm$0.00 & 0.49$\pm$0.32 & 0.81$\pm$0.03 & \omark \\
SMD-SSL + SimCLR (Frozen)                   & 0.84$\pm$0.01 & 0.83$\pm$0.00 & 0.15$\pm$0.05 & 0.61$\pm$0.02 & \omark  \\
SMD-SSL + VICReg (Frozen)                   & 0.79$\pm$0.01 & 0.79$\pm$0.00 & 0.12$\pm$0.05 & 0.60$\pm$0.03 & \omark  \\
SMD-SSL + SimCLR (Fine-tune)                & 0.94$\pm$0.01 & 0.94$\pm$0.00 & 0.39$\pm$0.10 & 0.78$\pm$0.04 & \omark  \\
SMD-SSL + VICReg (Fine-tune)                & 0.95$\pm$0.01 & 0.95$\pm$0.00 & 0.24$\pm$0.12 & 0.73$\pm$0.04 & \omark  \\       
\cmidrule(lr){1-6}
\textsc{Simple-SCL} / \textsc{Temporal-SCL} (PT:\xbmark, NN:\xbmark, TR:\xbmark)                                             & 0.93$\pm$0.01 & 0.94$\pm$0.01 & 0.19$\pm$0.01 & 0.64$\pm$0.06 & \xmark \\
\textsc{Temporal-SCL} (PT:\cbmark, NN:\xbmark, TR:\xbmark)      & 0.94$\pm$0.01 & 0.94$\pm$0.00 & 0.24$\pm$0.01 & 0.62$\pm$0.06 & \xmark \\
\textsc{Temporal-SCL} (PT:\xbmark, NN:\cbmark, TR:\xbmark)      & 0.94$\pm$0.01 & 0.95$\pm$0.00 & 0.82$\pm$0.01 & 0.84$\pm$0.05 & \omark \\
\textsc{Temporal-SCL} (PT:\xbmark, NN:\xbmark, TR:\cbmark)      & 0.95$\pm$0.01 & 0.95$\pm$0.00 & 0.40$\pm$0.01 & 0.57$\pm$0.04 & \xmark \\
\textsc{Temporal-SCL} (PT:\xbmark, NN:\cbmark, TR:\cbmark)      & 0.95$\pm$0.01 & 0.94$\pm$0.00 & 0.86$\pm$0.00 & 0.91$\pm$0.06 & \omark \\
\textsc{Temporal-SCL} (PT:\cbmark, NN:\xbmark, TR:\cbmark)      & 0.95$\pm$0.01 & 0.95$\pm$0.00 & 0.55$\pm$0.01 & 0.56$\pm$0.05 & \xmark \\
\textsc{Temporal-SCL} (PT:\cbmark, NN:\cbmark, TR:\xbmark)      & 0.95$\pm$0.01 & 0.95$\pm$0.00 & 0.92$\pm$0.01 & 0.85$\pm$0.03 & \omark \\
\textsc{Temporal-SCL} (Full)                                    & 0.95$\pm$0.01 & 0.95$\pm$0.00 & {\bftab 0.98$\pm$0.00} & {\bftab 1.00$\pm$0.00} & {\bftab \cmark}   \\ 
\bottomrule
\end{tabular}}

\vspace{.5ex}
{\scriptsize {\bftab{PT}}: Pre-Training, {\bftab{NN}}: Nearest Neighbor pairing, {\bftab{TR}}: Temporal Regularization}

\vspace{-.75ex}
{\scriptsize \xmark: No correct recovery of the embedding structure.}

\vspace{-.75ex}
{\scriptsize \omark: Partial clustering structure found, but not a completely correct recovery of the embedding structure.}

\vspace{-.75ex}
{\scriptsize \cmark: Completely correct recovery of the embedding structure.}
\label{tab:toyDataset_ratios}
\vspace{-1.5em}
\end{table*}

\vspace{.2em}
\noindent
\textbf{Experimental results.}
We use a 3D embedding space for all methods despite the ground truth being~2D.\footnote{Confining the embedding space to 2D markedly degraded performance across all methods. We speculate that this due to ``over-specification'', aligned with findings in studies like \citet{livni2014computational} that suggest that training larger-than-necessary neural networks results in easier training.}
Fig.~\ref{fig:toyDataset} depicts the test data projected onto a 3D embedding space of a few different methods (in the appendix, Fig.~\ref{fig:toyDataset_result_complete} has visualizations across all methods evaluated). 
We report test set accuracy scores for predicting red/blue (area under the ROC curve, area under precision recall curve), silhouette index (SI) using ground truth cluster centers as cluster labels, and lastly a qualitative analysis of whether all 10 clusters were recovered across the 10 experimental repeats (this is a qualitative visual check) in Table~\ref{tab:toyDataset_ratios}. 
Additionally, we also investigate a different downstream task focusing on cluster prediction accuracy. In this task, we freeze the encoder network, previously trained on the synthetic data, and task the model with predicting the cluster location for each timestep. This experiment is used to evaluate the model's ability to be used for different downstream tasks and evaluate whether the trained encoder can correctly identify the cluster of origin for each data point within the synthetic dataset's defined 10 clusters (as shown in Fig.~\ref{fig:toyDataset_Raw}). The cluster prediction accuracy is reported alongside other results in Table~\ref{tab:toyDataset_ratios}.

The main finding from our synthetic data experiment is that while most methods yield similar, optimal prediction scores, only the full version of \textsc{Temporal-SCL} consistently recovers all 10 clusters. This is evident in the high silhouette index (0.98) our full model achieves, closely matching the ground truth SI of 0.982. 
This claim is further supported by visual examination of Fig.~\ref{fig:toyDataset} (also see the last column of Table~\ref{tab:toyDataset_ratios} and appendix Fig.~\ref{fig:toyDataset_result_complete}). No other method evaluated consistently recovers all 10 ground truth clusters. Notably, removing nearest neighbor pairing from \textsc{Temporal-SCL} leads to a 0\% recovery rate of ground truth clusters. 
Lastly, our full model is the only one among all baselines and ablations to achieve 100\% accuracy in the downstream task of cluster prediction. This result highlights our model's ability to accurately identify cluster origins and provides empirical evidence of the versatility of the learned encoder representations for a different downstream task.

\vspace{-1em}   
\subsection{Real Clinical Data}
\label{sec:ClinicalExperiments}
\vspace{-.25em}

\begin{table*}[t]
\caption{Real data test set accuracy (mean $\pm$ std.~dev.~across 10 experiments).}
\centering
\adjustbox{max width=.8\linewidth}{%
\begin{tabular}{ccccc} 
\toprule
                        \multirow{2}{*}[-2pt]{Model}           & \multicolumn{2}{c}{MIMIC dataset}                  & \multicolumn{2}{c}{ADNI dataset}                                                   
                                                                        \\ \cmidrule(lr){2-3} \cmidrule(lr){4-5} 
                                                               & AUROC                      & AUPRC                   & AUROC                   & AUPRC \\ 
                        \midrule
                        Logistic Regression                    & 0.745$\pm$0.003            & 0.499$\pm$0.008         & 0.845$\pm$0.006          & 0.676$\pm$0.009 \\
					  LSTM                                   & 0.767$\pm$0.003            & 0.509$\pm$0.005         & 0.947$\pm$0.002          & 0.823$\pm$0.005 \\
                        RETAIN                                 & 0.730$\pm$0.010            & 0.431$\pm$0.006         & 0.884$\pm$0.012          & 0.795$\pm$0.016 \\
                        DIPOLE                                 & 0.767$\pm$0.004            & 0.453$\pm$0.003         & 0.958$\pm$0.006          & 0.824$\pm$0.009 \\
                        Transformer:BERT                       & 0.769$\pm$0.005            & 0.509$\pm$0.003         & 0.959$\pm$0.002          & 0.922$\pm$0.003 \\                        
                        AC-TPC                                 & 0.703$\pm$0.006            & 0.432$\pm$0.007         & 0.839$\pm$0.013          & 0.681$\pm$0.017 \\
                        T-Phenotype                            & 0.730$\pm$0.005            & 0.451$\pm$0.004         & 0.926$\pm$0.034          & 0.822$\pm$0.068 \\
                        SMD-SSL + SimCLR (Frozen)              & 0.692$\pm$0.002	        & 0.457$\pm$0.005	      & 0.803$\pm$0.015	         & 0.658$\pm$0.009 \\
                        SMD-SSL + VICReg (Frozen)              & 0.673$\pm$0.001	        & 0.429$\pm$0.005	      & 0.819$\pm$0.018	         & 0.653$\pm$0.009 \\
                        SMD-SSL + SimCLR (Fine-tune)           & 0.770$\pm$0.006	        & 0.511$\pm$0.010	      & 0.966$\pm$0.014	         & 0.929$\pm$0.010 \\
                        SMD-SSL + VICReg (Fine-tune)           & 0.748$\pm$0.008	        & 0.499$\pm$0.009	      & 0.867$\pm$0.012	         & 0.672$\pm$0.011 \\
                        \cmidrule(lr){1-5}
\textsc{Simple-SCL} / \textsc{Temporal-SCL} (PT:\xbmark, NN:\xbmark, TR:\xbmark)                                          & 0.753$\pm$0.005            & 0.498$\pm$0.003         & 0.947$\pm$0.001          & 0.894$\pm$0.015 \\
\textsc{Temporal-SCL} (PT:\cbmark, NN:\xbmark, TR:\xbmark)    & 0.752$\pm$0.002            & 0.498$\pm$0.001         & 0.947$\pm$0.001          & 0.894$\pm$0.014 \\
\textsc{Temporal-SCL} (PT:\xbmark, NN:\cbmark, TR:\xbmark)    & 0.770$\pm$0.002            & 0.516$\pm$0.003         & 0.987$\pm$0.001          & 0.899$\pm$0.011 \\
\textsc{Temporal-SCL} (PT:\xbmark, NN:\xbmark, TR:\cbmark)    & 0.766$\pm$0.001            & 0.508$\pm$0.002         & 0.950$\pm$0.002          & 0.785$\pm$0.015 \\
\textsc{Temporal-SCL} (PT:\xbmark, NN:\cbmark, TR:\cbmark)    & 0.754$\pm$0.006            & 0.499$\pm$0.003         & 0.967$\pm$0.002          & 0.900$\pm$0.017 \\ 
\textsc{Temporal-SCL} (PT:\cbmark, NN:\xbmark, TR:\cbmark)    & 0.767$\pm$0.002            & 0.511$\pm$0.002         & 0.951$\pm$0.002          & 0.762$\pm$0.010 \\
\textsc{Temporal-SCL} (PT:\cbmark, NN:\cbmark, TR:\xbmark)    & 0.770$\pm$0.002            & 0.518$\pm$0.003         & 0.988$\pm$0.001          & 0.903$\pm$0.011 \\
                        \textsc{Temporal-SCL} (Full)          & {\bftab 0.773$\pm$0.002}   &{\bftab 0.520$\pm$0.003} & {\bftab 0.990$\pm$0.004} & {\bftab 0.936$\pm$0.014} \\ 
\bottomrule
\end{tabular}}

\vspace{.5ex}
{\scriptsize {\bftab{PT}}: Pre-Training, {\bftab{NN}}: Nearest Neighbor pairing, {\bftab{TR}}: Temporal Regularization\\}
\label{tab:results}
\vspace{-1em}
\end{table*}

\noindent
\textbf{Data.}
We use two standard datasets: 
\begin{itemize}[leftmargin=*,itemsep=0pt,parsep=0pt,topsep=1pt]
\item[$\bullet$]
\textbf{MIMIC }(static outcome case)\textbf{.}
We employ time series data of septic patients from the MIMIC-III dataset (v1.4) \citep{johnson2018mimic} following the same procedure as \citet{komorowski2018artificial} to identify 18,354 patients with sepsis onset based on Sepsis-3 criteria. 
The prediction task among this sepsis cohort is ICU visit mortality, where only the final outcome label is known 
 ($y_i^{(L_i)}$) determined by death within 48h of the final observation $(=1)$ or death within 90 days of the final observation $(=2)$ versus discharge $(=0)$.

\item[$\bullet$]
\textbf{ADNI }(dynamic outcome case)\textbf{.} 
We also evaluate our approach using the Alzheimer's Disease Neuroimaging Initiative (ADNI) dataset \citep{petersen2010alzheimer}, encompassing 11,651 hospital visits across 1,346 patients, tracking AD progression through 6-month interval follow-ups.
Our prediction task mirrors \cite{lee2020outcome} for multiclass classification of every time step ($y_i^{l}$) among three diagnostic groups: normal brain function $(=0)$, mild cognitive impairment $(=1)$, and AD $(=2)$.
This setup covers the dynamic outcome case where the label at both terminal and non-terminal time steps are known.
We further predict cognitive impairment status 6 months and 1 year in advance in Appendix~\ref{sec:future_adni}.

\end{itemize}
More details on the datasets and how we preprocess them (including imputation) are in Appendix~\ref{sec:real-data-info}.

\noindent
\textbf{Experimental results.} Accuracy scores for methods evaluated are in Table \ref{tab:results}. In both MIMIC and ADNI datasets, \textsc{Temporal-SCL} (the full model without ablation) yields strong results, outperforming all baseline approaches. Omitting pre-training from \textsc{Temporal-SCL} substantially decreases accuracy. Meanwhile, excluding nearest neighbor pairing shows accuracy closer to the full model in MIMIC but significantly lower in ADNI. We suspect that all models typically work better in ADNI vs MIMIC due to the ADNI dataset having less missing data and less feature noise (e.g., ADNI has memory assessments as features whereas MIMIC has lab exams, which tend to be noisier). 

We expand our analysis of the heatmap visualization in Section~\ref{sec:visualizing} for the complete \textsc{Temporal-SCL} model trained on MIMIC. A more comprehensive version of Fig.~\ref{fig:heatmapMIMIC}, encompassing all features, is in Appendix~\ref{sec:heatmap_mimic}, with clustering details outlined in Appendix~\ref{sec:cluster}. 
For the sepsis cohort in MIMIC, our visualization reveals that higher AST, lactate, ALT, and INR levels appear associated with higher mortality risk. Moreover, lower bicarbonate levels also appear associated with higher mortality risk.
These model-derived insights agree with clinical literature on sepsis \citep{nesseler2012clinical, villar2019lactate}, where abnormal biomarker ranges correlate with heightened sepsis-related mortality.

We also applied our heatmap visualization to the \textsc{Temporal-SCL} model for ADNI. The full heatmap and analysis are in Appendix~\ref{sec:heatmap_adni}. In brief, for ADNI, \textsc{Temporal-SCL} reveals a correlation between low Rey Auditory Verbal Learning Test (RAVLT) Immediate scores, high RAVLT Forgetting scores, and high Clinical Dementia Rating – Sum of Boxes (CDR-SB) values with increased Alzheimer's disease risk. We cross-verified these results with Alzheimer's disease literature \citep{o2008staging, moradi2017rey}, confirming our model's captured correlations.

Importantly, our heatmap visualizations can show when two clusters have similar outcomes (e.g., similar in-hospital mortality rates in the case of MIMIC data) but different patient characteristics.
For example, in Fig.~\ref{fig:heatmapMIMIC}, clusters with mortality risks of $0.737$ and $0.740$ showcase distinct patient traits within physiological features, despite similar mortality rates. Similarly, the clusters with mortality risks of $0.492$ and $0.523$ also have different patient characteristics.

\vspace{.2em}
\noindent
\textbf{Ablation studies.}
While we have already pointed out the importance of pre-training and of nearest neighbor pairing in \textsc{Temporal-SCL}, we provide more detailed ablation experiment results on both synthetic and the real clinical datasets in Appendix~\ref{sec:ablation}.

\vspace{-1em} 
\section{Discussion}
\vspace{-.25em}
We have proposed a variable-length tabular time series framework called \textsc{Temporal-SCL} that outperforms state-of-the-art models on two clinical datasets and on a synthetic dataset with known ground truth structure. A key ingredient to the success of \textsc{Temporal-SCL} is the nearest neighbor pairing mechanism. 
We further suggested a visualization strategy to help probe \textsc{Temporal-SCL}'s embedding space.

We highlight some limitations of our work that in turn suggest future research directions. First, we make predictions only for single time steps. Extending our framework to make predictions for an entire time series would make it more useful as it can use the entire information available at prediction time. Next, while we have demonstrated that nearest neighbor pairing works well, we do not understand when and why. For future work, we could empirically try different distance functions that can be used to answer the former question and also study an unsupervised \textsc{Temporal-SCL} variant (nearest neighbor pairing could still be used by ignoring labels) or \textsc{Temporal-SCL} with other label information (e.g., survival analysis labels). Lastly, our proposed strategy for visualizing the embedding space also focuses on individual time steps. Figuring out a visualization strategy that considers entire trajectories, even variable-length ones, could be useful for clinical decision support.

\vspace{-1.25em}
\section*{Acknowledgments}
\vspace{-.25em}
This research was supported in part by the Intramural Research Program of the National Library of Medicine (NLM), National Institutes of Health (NIH). 
S.~N.~was supported by a fellowship from Carnegie Mellon University's Center for Machine Learning and Health. G.~H.~C.~was supported by NSF CAREER award \#2047981. The authors thank Rema Padman and Zachary Lipton for helpful discussions.

\bibliography{noroozizadeh23}

\newpage
\appendix
\section{Appendix}\label{apd:first}
This appendix is structured as follows. 
In Section~\ref{sec:relating-to-original-SCL}, we compare the Simple-SCL method with the original Supervised Contrastive Learning (SCL), highlighting the use of data augmentation in SCL. Section~\ref{sec:handling-static-outcome} outlines adaptations for static outcome cases in Temporal-SCL, focusing on encoder pre-training and joint training modifications. In Section~\ref{sec:data-augmentation-with-nn-pairing}, we discuss how in our model we can integrate data augmentation with nearest neighbor pairing. 

In Section~\ref{sec:experiment-details}, we delve into the details of our experiments and our model setup. Section~\ref{sec:baselines} provides an overview of the baselines used in our experiments, categorized into purely-supervised, predictive clustering, and self-supervised learning methods. 
Detailed experimental setup, including training hyperparameters and evaluation metrics, are presented in Section~\ref{sec:ExperimentalSetup}, while our synthetic data generation is outlined in Section~\ref{sec:syntheticData_full} and our specific approaches for real-world datasets like MIMIC and ADNI, and data imputation strategies are discussed in Section~\ref{sec:real-data-info}. Section \ref{sec:cluster} discusses the strategy for clustering on the embedding space for visualization. 

Finally, Section~\ref{sec:experimental-results-details} delves into the details of the results from synthetic and real data experiments.
In Section~\ref{sec:ToyExperiment_Results}, we present more comprehensive results from our synthetic data experiments. Sections~\ref{sec:heatmap_mimic} and \ref{sec:heatmap_adni} focus on visualizing embedding clusters for MIMIC and ADNI datasets, respectively. Section~\ref{sec:heatmap_mimic_raw} compares these visualizations with those derived from raw features. Lastly, in Section~\ref{sec:ablation}, we assesse the impact of different model components and in Section~\ref{sec:future_adni} we present an additional experiment on future cognitive impairment prediction in the ADNI dataset.

\subsection{\textsc{Simple-SCL} vs the Original SCL}
\label{sec:relating-to-original-SCL}
The original SCL uses data augmentation: in the loss function $L_{\text{SCL}}$, instead of using the batch of $B$ points $(\mathbf{x}_1,y_1),\dots,(\mathbf{x}_B,y_B)$, we instead use a freshly generated batch of $2B$ points $(\mathbf{x}_1',y_1'),\dots,(\mathbf{x}_{2B}',y_{2B}')$. Specifically, each original data point $\mathbf{x}_i$ is randomly augmented once to get $\mathbf{x}_{2i-1}'$ and then $\mathbf{x}_i$ is randomly augmented a second time to get $\mathbf{x}_{2i}'$; the augmented points $\mathbf{x}_{2i-1}'$ and $\mathbf{x}_{2i}'$ have labels $y_{2i-1}'$ and $y_{2i}'$, both set to be the same as~$y_i$.

\subsection{Handling the Static Outcome Case}
\label{sec:handling-static-outcome}

To train a \textsc{Temporal-SCL} model for the static outcome case described in Section~\ref{sec:problem-setup}, there are some small modifications to what we described for pre-training the encoder (Section~\ref{sec:learning-the-encoder}) and the joint encoder and temporal network training (Section~\ref{sec:joint-optimization}).

\vspace{.2em}
\noindent
\textbf{Pre-training the encoder.}
In the dynamic outcome case (i.e., the classification outcome changes over time), we use the procedure stated in Section~\ref{sec:learning-the-encoder}. However, in the static outcome case (i.e., the classification label is for the final time step only), during the pre-training phase, we only use the snapshot corresponding to the final time step per training time series. This is because we are not sure of what the true labels should be prior to the final time step, so for the pre-training we focus learning the embedding vectors based on time steps where we know the labels.

\vspace{.2em}
\noindent
\textbf{Encoder and temporal network training.} Once again, in the dynamic outcome case, we use the procedure stated in Section~\ref{sec:joint-optimization}. However, in the static outcome case, for the joint optimization phase, we use all snapshots (unlike during pre-training). As a reminder, snapshots that do not correspond to the final time step of a time series has the classification label ``?'' which as stated in Section~\ref{sec:background} could be thought of as an additional ``unknown'' class. The idea is that now that we are accounting for temporal structure, despite us not knowing the labels prior to the final time step, we still want to encourage temporal smoothness of embedding vectors across time steps.

\subsection{Combining Data Augmentation with Nearest Neighbor Pairing}
\label{sec:data-augmentation-with-nn-pairing}

When data augmentation is available, nearest neighbor pairing can be run using an augmented training dataset. We keep track of what the original $N$ raw time series are prior to any data augmentation. Per training epoch, we use a different random augmentation of every original training time series and treat the augmented set of $N$ time series as a ``fresh'' set of training feature vectors for that epoch (prior to running our nearest neighbor pairing procedure).

Alternatively, we can adopt the approach used in the original SCL, which involves pairing two random augmentations of the same raw input. This would require generating two random augmentations per training time series in each epoch. In fact, combining these two strategies for contrastive learning on images has been previously done \citep{dwibedi2021little}; however, this earlier work finds nearest neighbors in the embedding space rather than the raw feature space and has not been extended to variable-length irregularly sampled temporal data like the data we experiment on.

\vspace{-.25em}
\subsection{More Details on Experiments}
\label{sec:experiment-details}

\subsubsection{Baselines}
\label{sec:baselines}
In this section we describe the baselines used for our experiments. We divide these baselines into 3 subgroups of: (a) Purely-supervised, (b) Predictive clustering, and (c) Self-supervised learning and contrastive learning methods.

\begin{itemize}
    \item \textbf{Purely-Supervised Baselines}
    \vspace{-0.5em}
\end{itemize}
\noindent
\textbf{Logistic regression.}
For this simple baseline, our training data has each timestep as an individual datapoint with the true label of each timestep being the final timestep outcome of its corresponding sequence. 
We run logistic regression on raw features of every timestep in the training set.

\vspace{.2em}
\noindent
\textbf{Long short-term memory (LSTM).}
We train a multilayer LSTM \citep{hochreiter1997long} on the sequences of the raw features in the training set and their corresponding outcome label and take the hidden layer of the final timestep and pass it through a linear layer to predict the outcome of the corresponding sequence.
After the training is completed, we run our model on the validation set and get the probability of the positive outcome (not surviving) for each sequence.

\vspace{.2em}
\noindent
\textbf{Reverse Time Attention Model (RETAIN).}
\citet{choi2016retain} proposed an interpretable predictive model for healthcare using reverse time attention mechanism. Their proposed model is based on a two-level neural attention model. It detects influential past visits and significant clinical variables within those visits by mimicking physician practice. The model attends to the EHR data in a reverse time order, giving higher attention to recent clinical measurements.

\vspace{.2em}
\noindent
\textbf{Diagnosis prediction in healthcare via attention-based bidirectional recurrent neural networks (DIPOLE).}
\citet{ma2017dipole} propose a model that utilizes bidirectional recurrent neural networks to remember all the information of both the past visits and the future visits. The authors introduce three attention mechanisms to measure the relationships of different visits for the prediction. Dipole uses the attention mechanism to interpret the prediction results it generates.

\vspace{.2em}
\noindent
\textbf{Transformer: Bidirectional Encoder Representations from Transformers (BERT).} 
We train a BERT-based \citep{devlin2018bert} model from scratch to encode the time-series data of our experiments to predict the final outcome of each time-series.

\begin{itemize}
    \item \textbf{Predictive-Clustering Baselines}
    \vspace{-0.5em}
\end{itemize}
\noindent
\textbf{Actor-Critic Temporal Predictive Clustering (AC-TPC).}
\citet{lee2020temporal} proposed an actor-critic approach from reinforcement learning for temporal predictive clustering (AC-TPC) where each cluster consists of patients with similar future outcomes of interest.
In this approach an RNN-based encoder and multi-layer perceptron predictor network are first trained to initialize time series embeddings.
After initialization, a selector network and an embedding dictionary are jointly optimized with the encoder and predictor networks to obtain a representative embedding for each timestep, considering the outcome in the subsequent timestep. During inference at each timestep, the encoder maps a sequence into a latent embedding. Subsequently, the selector network assigns a cluster to this latent embedding. The centroid of the cluster assigned to the latent embedding is stored in the embedding dictionary and used by the predictor network to make predictions regarding the future outcome of interest.

\vspace{.2em}
\noindent
\textbf{T-Phenotype.}
\citet{qin2023t} proposed a method for temporal clustering designed to uncover predictive temporal patterns from labeled time-series data, aiding in understanding disease progression. This method focuses on processing multivariate time-series data to extract predictive patterns and associate them with relevant clinical progression markers. At the heart of T-Phenotype is the utilization of the Laplace transform for representation learning, which transforms variable-length, irregularly sampled time-series data into a consistent and unified embedding in the frequency domain. This transformation is crucial for handling the inherent variability and complexity of clinical data. The model further includes a path-based similarity score, a novel metric to assess the relationship between temporal patterns and specific disease statuses. This score is instrumental in determining the relevance and predictive power of identified patterns. T-Phenotype then employs a graph-constrained K-means clustering process, guided by the path-based similarity graph, to categorize patients into distinct phenotypes. These phenotypes are defined by their unique temporal patterns, providing insights into different trajectories of disease progression and offering potential for personalized patient care strategies.

\begin{itemize}
    \item \textbf{Self-Supervised Learning and Contrastive Learning Baselines}
    \vspace{-0.5em}
\end{itemize}
\noindent
\textbf{Sequential Multi-Dimensional Self-Supervised Learning (SMD-SSL).}
\citet{raghu2023sequential} developed Sequential Multi-Dimensional Self-Supervised Learning (SMD-SSL) to effectively process complex clinical time series data, particularly in settings with abundant multimodal data like intensive care units. This method addresses the shortcomings of traditional self-supervised learning approaches that are limited to unimodal data. SMD-SSL is capable of handling both structured data, such as lab values and vital signs, and high-dimensional physiological signals like electrocardiograms. It employs a dual SSL loss optimization strategy, applying one loss at the level of individual high-dimensional data points and another at the overall sequence level, enabling a more nuanced understanding of the multimodal clinical data. Crucially, SMD-SSL is designed to be compatible with different loss functions, allowing the use of either contrastive methods, as exemplified by SimCLR \citep{chen2020simple}, or non-contrastive approaches like VICReg \citep{bardes2021vicreg}. In our experiments, we implement SMD-SSL using both the contrastive loss from SimCLR and the non-contrastive loss from VICReg. Following the self-supervised learning phase with each of these loss functions, we attach a prediction head to the model for our prediction tasks. We present two variations of the model in our experiments: a fine-tuned version, where the loss is back-propagated through both the prediction head and the main SMD-SSL encoder, and a frozen version, in which we keep the main SMD-SSL encoder weights constant and optimize only the prediction head.

\vspace{.2em}
\noindent
\textbf{\textsc{Simple}-Supervised Contrastive Learning ({\textsc{Simple-SCL}})}
This baseline lies under the umbrella of embedding-centric models. \textsc{Simple-SCL} is a special case of the original \textsc{SCL} \citep{khosla2020supervised} which does not use data augmentation. Similar to the original \textsc{SCL} it includes both an encoder network that first learns the embeddings and then a predictor network is trained on these learned embeddings to make predictions. In our experiments that deals with temporal data, we treat each time step as an individual data point to train the encoder and predictor networks. \emph{Note that \textsc{Simple-SCL} can be viewed as an ablation of our full \textsc{Temporal-SCL} model that excludes both the temporal network and the nearest neighbor pairing mechanism.}  

\subsubsection{Experimental Setup Details}
\label{sec:ExperimentalSetup}

For all datasets, we randomly split the data into 60\% training, 20\% validation, and 20\% test sets. 
This split is done at the patient level. The presented results of our experiments in the following sections would be for holdout test patients from which no training data was available. We train our model on the patients in the training set, optimize the hyperparameters on the patients in the validation set and report the results from the holdout patient test set. 
We train each method on the training set, tune hyperparameters based on the validation set, and report evaluation metrics on the test set.
Furthermore, we choose 10 different random seeds to randomize the parameter initialization of all the models evaluated as well as randomizing the train/validation/test sets for each experimental repeat.

\vspace{.2em}
\noindent
\textbf{Evaluation metrics.}
We use the same evaluation metrics as \citet{lee2020temporal}. To evaluate the supervised performance of our model and the baselines, we use area under receiver operator characteristic curve (AUROC) and area under precision-recall curve (AUPRC) obtained from the label predictions of our model and the ground-truth labels on the outcomes of interest. We use one-vs-rest AUROC and AUPRC. For the synthetic data experiment, we also look at the fraction of experimental repeats that a method recovers the ground truth structure. We also report the Silhouette index \citep{rousseeuw1987silhouettes} that gives us a measure of how similar a member is to its own cluster (homogeneity within a cluster) compared to other clusters (heterogeneity across clusters). 

In the ablation studies presented in Section~\ref{sec:ablation}, we also evaluate the unsupervised performance of different ablations of \textsc{Temporal-SCL}.
We cluster on the learned hyperspherical embeddings using a complete linkage Agglomerative Hierarchical Clustering to discover discrete latent states that could be of interest.  
We utilize three standard metrics for the scenario when ground-truth label is known and a fourth completely unsupervised metric.
The metrics used are purity score, normalized mutual information (NMI), adjusted rand index (ARI), and silhouette index (SI).
Purity score ranges from 0 to 1 and explains the homogeneity of each cluster with regards to the labels.
NMI ranges from 0 to 1 and is an information theoretic measure of mutual information shared between the labels and the cluster and is adjusted for the number of clusters with 1 being perfect clustering.
ARI ranges from -1 to 1 and evaluates the percentage of correct cluster assignments where 0 corresponds to completely random assignment and 1 being perfect clustering.
Lastly, SI is a completely unsupervised metric ranging from -1 to 1 which provides a simultaneous measure of (1) how similar members of a cluster are to their own cluster capturing the homogeneity within a cluster and (2) how different members of each cluster are compared to other clusters capturing heterogeneity across different clusters.

\vspace{.2em}
\noindent
\textbf{Training hyperparameters.}
Table~\ref{tab:trainingHyperParams} summarizes the hyperparameters used in all our experimentations presented in this paper.
{
\begin{table*}[!ht]
\caption{Training Details and Hyperparamters for Experiments.}
\centering
\begin{tabular}{|l|l|}
\hline
\textbf{Experiment}                    & \textbf{Synthetic Dataset}                                                                                                             \\ \hline
\textit{Main Network}                  & 2-Layer FCN ReLU Activation                                                                                                            \\
\textit{Main Network Layer Size}       & (input-dimension)2→16→3(embedding-space)                                                                                               \\
\textit{Prediction Network}              & Linear Layer with Softmax Activation 3→2                                                                                               \\
\textit{Temporal Module}               & 1-Layer LSTM 3→3                                                                                                                       \\
\multirow{2}{*}{\textit{Optimization}} & \multirow{2}{*}{\begin{tabular}[c]{@{}l@{}}Optimizer: Adam, Learning Rate: 1e-4, \\ Number of Epochs: 20, Batch-Size: 32\end{tabular}} \\
                                       &                                                                                                                                        \\ \hline
\textbf{Experiment}                    & \textbf{MIMIC}                                                                                                                         \\ \hline
\textit{Main Network}                  & 2-Layer FCN ReLU Activation                                                                                                            \\
\textit{Main Network Layer Size}       & (input-dimension)55→32→32(embedding-space)                                                                                             \\
\textit{Prediction Network}              & Linear Layer with Softmax Activation 32→3                                                                                              \\
\textit{Temporal Module}               & 1-Layer LSTM 32→32                                                                                                                     \\
\textit{Optimization}                  & \begin{tabular}[c]{@{}l@{}}Optimizer: Adam, Learning Rate: 1e-4, \\ Number of Epochs: 100, Batch-Size: 128\end{tabular}                \\ \hline
\textbf{Experiment}                    & \textbf{ADNI}                                                                                                                          \\ \hline
\textit{Main Network}                  & 2-Layer FCN ReLU Activation                                                                                                            \\
\textit{Main Network Layer Size}       & (input-dimension)21→50→16(embedding-space)                                                                                             \\
\textit{Prediction Network}              & Linear Layer with Softmax Activation 16→3                                                                                              \\
\textit{Temporal Module}               & 1-Layer LSTM 16→16                                                                                                                     \\
\textit{Optimization}                  & \begin{tabular}[c]{@{}l@{}}Optimizer: Adam, Learning Rate: 1e-4, \\ Number of Epochs: 100, Batch-Size: 128\end{tabular}                \\ \hline
\end{tabular}
\label{tab:trainingHyperParams}
\end{table*}
}

\subsubsection{Synthetic Data}
\label{sec:syntheticData_full}
We generate a 2D dataset where every time series has exactly 3 time steps. We consider the static outcome case so that each time series has a single label (one of two classes: red or blue). The points are all on a 2D circle, where the only four possible time series in the embedding space are shown in Figure~\ref{fig:toyDataset_Raw} and  include: (i) ``\scalerel*{\includegraphics{Figures/circle-blue}}{B} $\rightarrow$ \scalerel*{\includegraphics{Figures/triangle-down-blue}}{B} $\rightarrow$ \scalerel*{\includegraphics{Figures/star-blue}}{B}'' and (ii) ``\scalerel*{\includegraphics{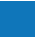}}{B} $\rightarrow$ \scalerel*{\includegraphics{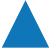}}{B} $\rightarrow$ \scalerel*{\includegraphics{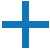}}{B}'' for the \textcolor{blue}{blue} class and iii) ``\scalerel*{\includegraphics{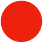}}{B} $\rightarrow$ \scalerel*{\includegraphics{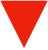}}{B} $\rightarrow$ \scalerel*{\includegraphics{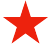}}{B}'' and (iv) ``\scalerel*{\includegraphics{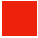}}{B} $\rightarrow$ \scalerel*{\includegraphics{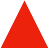}}{B} $\rightarrow$ \scalerel*{\includegraphics{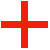}}{B}'' for the \textcolor{red}{red} class. We place the starting time steps (``\scalerel*{\includegraphics{Figures/circle-blue}}{B}'', ``\scalerel*{\includegraphics{Figures/square-blue}}{B}'', ``\scalerel*{\includegraphics{Figures/circle-red}}{B}'', ``\scalerel*{\includegraphics{Figures/square-red}}{B}'') on angular points $\{0^{\circ}, 180^{\circ}, 0^{\circ}, 180^{\circ}\}$ of the circle circumference as shown on Figure~\ref{fig:toyDataset_Raw}. For these starting points we purposefully have the starting points of the two different classes fall in the same region.
Next for the second time steps (``\scalerel*{\includegraphics{Figures/triangle-down-blue}}{B}'', ``\scalerel*{\includegraphics{Figures/triangle-up-blue}}{B}'', ``\scalerel*{\includegraphics{Figures/triangle-down-red}}{B}'', ``\scalerel*{\includegraphics{Figures/triangle-up-red}}{B}''), we place them on the angular points of $\{45^{\circ}, 135^{\circ}, -45^{\circ}, -135^{\circ}\}$. And finally for the terminal time steps (``\scalerel*{\includegraphics{Figures/star-blue}}{B}'', ``\scalerel*{\includegraphics{Figures/plus-blue}}{B}'', ``\scalerel*{\includegraphics{Figures/star-red}}{B}'', ``\scalerel*{\includegraphics{Figures/plus-red}}{B}''), we place them on angular points of  $\{80^{\circ}, 100^{\circ}, -80^{\circ} -100^{\circ}\}$.
Given the clashing of the two starting time steps, there are a total of 10 true embedding locations (which could be thought of as cluster centers). When we generate synthetic time series, each point is based on one of the 10 true ground truth embedding locations with Gaussian noise ($\mathcal{N}(0,8)$) added to the true angles noted above. We randomly sample 200 of each of the 4 possible 3-time stepped trajectories so that we have a total of 800 time series.

\begin{figure*}[!ht]
\vspace{-.25em}
\centering
\includegraphics[width=.99\linewidth]{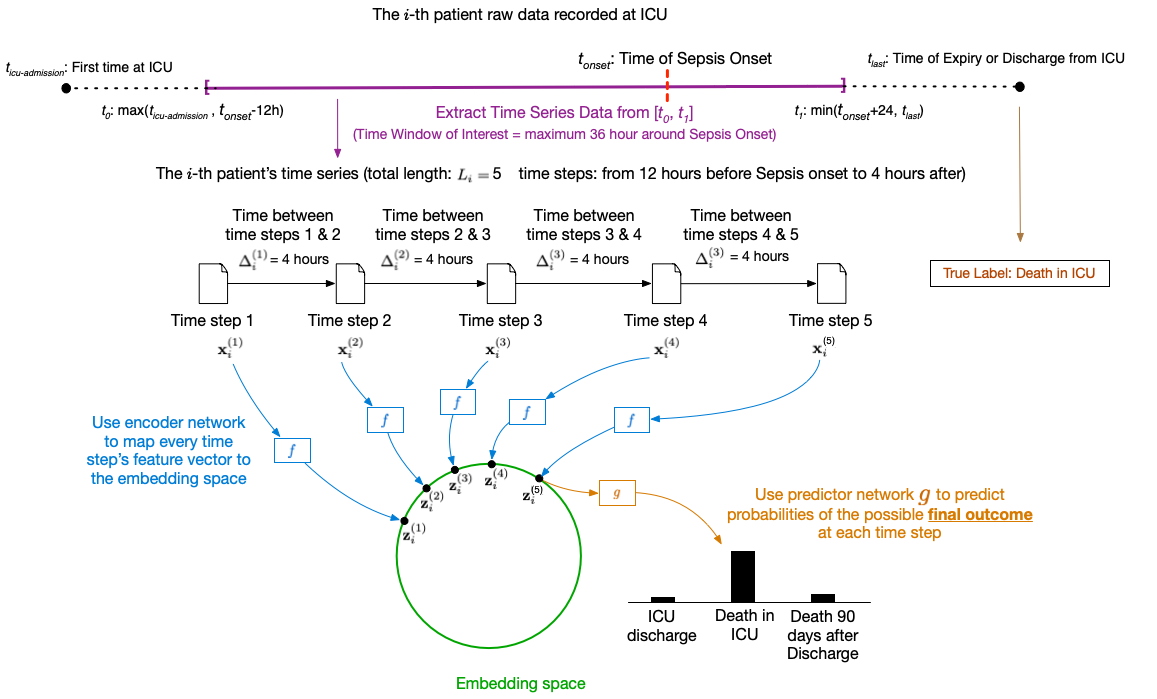}
 \caption{Setup for MIMIC Experiments at Inference time from \textsc{Temporal-SCL}: First, we extract time series data for each patient from a time window around their sepsis onset within their complete ICU data timeline. This time series data is transformed into multiple time steps at 4-hour intervals. The resulting time series will have its features at each time step mapped onto the embedding space learned by our encoder. Finally, these resulting embeddings will be passed through our predictor network to predict the ICU mortality of each patient at every time step.
 }   \label{fig:mimic-setup}
\end{figure*}

\begin{figure*}[!ht]
\vspace{-.25em}
\centering
\includegraphics[width=.99\linewidth]{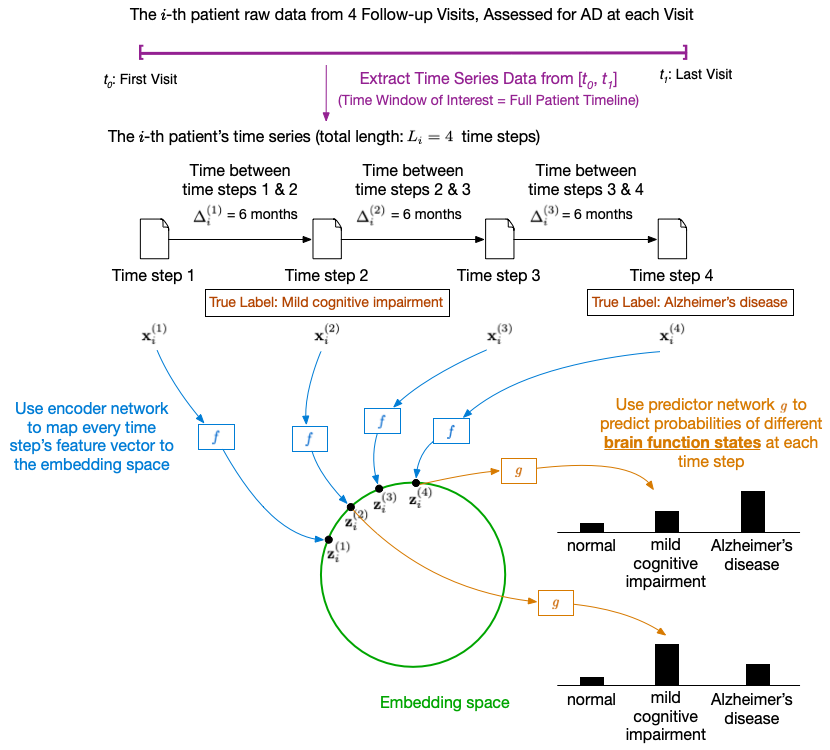}
 \caption{Setup for ADNI Experiment at Inference time from \textsc{Temporal-SCL}: First, we extract time series data for each patient from their complete timeline of available data spanning all the 6-month follow-up visits. This time series data maintains the same 6-month interval time steps as the raw data, with each time step having its own true class label representing one of the three possible brain function states. The resulting time series will have its features at each time step mapped onto the embedding space learned by our encoder. Subsequently, these resulting embeddings will be passed through our predictor network to predict the brain function class of each patient at every time step.
 }   \label{fig:adni-setup}
\end{figure*}

\subsubsection{Real Data}
\label{sec:real-data-info}

\textbf{Medical Information Mart for Intensive Care (MIMIC).}
We consider the trajectory of septic patients using data from the Medical Information Mart
for Intensive Care (MIMIC-III) dataset (v1.4) \citep{johnson2018mimic}. We follow the same procedure as done by \citet{komorowski2018artificial} to identify 18,354 septic patients among which there is an observed mortality rate just above 20\% (determined by death within 48h of the final observation $y_i^{(L_i)}=1$ or death within 90 days of the final observation $y_i^{(L_i)}=2$ versus discharge $y_i^{(L_i)}=0$).
From MIMIC, we extract demographic, lab results, and physiological features according to \citet{seymour2019derivation}, resulting in 29 features. Note that Seymour et al.~group these features as follows: (1) Hepatic: Bilirubin, AST, ALT; (2) Hematologic: Hemoglobin, INR, Platelets; (3) Neurologic: GCS; (4) Cardiovascular: Heart rate, Systolic blood pressure, Bicarbonate, Troponin, Lactate; (5) Pulmonary: Respiratory rate, SaO2, PaO2; (6) Inflammatory: Temperature, ESR, WBC count, Bands, C-Reactive protein; (7) Renal: Serum creatinine; (8) Other: Age, Gender, Elixhauser, Albumin, Chloride, Sodium, Glucose, BUN. 
We compile the measurements recorded of these features in MIMIC for every 4 hours (duration of each timestep). 
Our temporal data extraction follows the exact experimental procedure as \citet{komorowski2018artificial} that is also adopted by \citet{killian2020empirical}. Specifically, in the procedure that we followed, for the septic patients identified, we first found the time for the onset of sepsis based on Sepsis3 Criteria as $t_{\text{onset}}$, we then gathered recorded data of our selected features from the time 
$t_0 = \max(t_{\text{icu-admission}}, t_{\text{onset}}-12)$ to the time $t_1 = \min(t_{\text{icu-discharge}} \text{ or } t_{\text{death}}, t_{\text{onset}}+24)$. This results in a maximum timeline $[t_0, t_1]$ of 36 hours worth of information captured from each patient (i.e. the maximum number of time steps would be 10 for a single patient) from $-12$ hours of the onset of sepsis to $+24$ hours afterwards.
We impute the missing values from the population median if a measurement is not recorded before and the previously recorded value if the measurement is recorded in a previous timestep. 
We also include a set of 26 indicators for our time-varying features at each timestep that tracks whether a measurement was recorded ($=1$) or imputed ($=0$).
This dataset represent the case where only the final outcome-label is known and non-terminal timesteps have unknown state-label.
Figure~\ref{fig:mimic-setup} shows the setup for our MIMIC-Sepsis experiment.

\vspace{.2em}
\noindent
\textbf{Alzheimer's Disease Neuroimaging Initiative (ADNI).}
We also test our method on the Alzheimer's Disease Neuroimaging Initiative (ADNI) dataset \citep{petersen2010alzheimer}. 
This dataset consists of a total of 11,651 hospital visits from 1,346 patients which tracks the progression of Alzheimer's disease via follow-up observations at 6 months interval. 
Each patient has 21 variables out of which 5 are static and 16 are time-varying. 
The features include information on demographics, biomarkers of the brain function, and cognitive test results. 
These static features are as follows: (1) Demographic: Race, Ethnicity, Education, Marital Status; (2) Genetic: APOE$_4$.
The time-varying features include: (3) Demographic: Age, (4) Biomarker: Entorhinal, Fusiform, Hippocampus, Intracranial, Mid Temp, Ventricles, Whole Brain; and (5) Cognitive: ADAS-11, ADAS-13, Clinical Dimentia Rating Sum of Boxes (CDR-SB), Mini Mental State, Rey's Auditory Verbal Learning Test scores (RAVLT) Forgetting, RAVLT Immediate, RAVLT Learning, RAVLT Percent.
Following \citet{lee2020temporal}, we set our predictions on three diagnostic groups of normal brain functioning ($y_i^{(\ell)}=0$), mild cognitive impairment ($y_i^{(\ell)}=1$), and Alzheimer’s disease ($y_i^{(\ell)}=2$). 
The diagnostic is known at every timestep.
This dataset represent the case where we know the outcome-label at both terminal and non-terminal timesteps. 
Figure~\ref{fig:adni-setup} shows the overview for our ADNI experiment that follows \citet{lee2020temporal} setup.

\vspace{.2em}
\noindent
\textbf{Imputation.}
For both real datasets of this paper, we take the following approach to impute the missing features in our experiments. 
(1) For each patient, if at any time step feature has been recorded in a previous time step, we use the last recorded value to replace the missing feature.
(2) If at no previous time step a feature is recorded, we calculate the population median of the feature among all patients and impute the missing feature with this value.
In the future work, we aim to extend our approach to be similar to that of \citet{seymour2019derivation}, where multiple imputation with chained equations was used to account for missing data. 

In our MIMIC experiments, for each timestep, we include 26 indicators tracking measurement recording ($=1$) or imputation ($=0$) for time-varying features.

\subsubsection{Clustering on the Embedding Space for Our Visualization Strategy}
\label{sec:cluster}
For clustering on the embedding space, we use complete linkage Agglomerative Hierarchical Clustering. 
We first obtain the embedding representation of our training dataset and then train the clustering algorithm for $K\in\{5,7,8,9,10,14,20\}$. 
Subsequently, for predictive cluster assignment, we train a 3-Nearest Neighbor classifier (KNN) on the training data, utilizing the cluster assignments obtained in the previous step.
We use our classifier to get the predicted cluster assignment for the validation set.

\subsection{Experimental Results Details}
\label{sec:experimental-results-details}
\subsubsection{Synthetic Data Results}
\label{sec:ToyExperiment_Results}

In Figure~\ref{fig:toyDataset_result_complete}, we visualize the embedding space of all the models tested. For all the baselines tested, we show the 3D embedding space. For the \textsc{Temporal-SCL} model and its ablations where the embedding is constrained to lie on the hypersphere, we also plot the underlying unit hypersphere shell as well as the embeddings.

\begin{figure*}[htbp]
\floatconts
    {fig:toyDataset_result_complete}
    {\caption{Embedding space representation of the test trajectories of simulated dataset. For (a)- (j) 3D Embedding Space of baselines are shown which includes the purely supervised methods ((a)-(d)), predictive clustering methods ((e), (f)), and self-supervised learning methods ((g)-(j)). In (f)-(i), Hyperspherical Embedding of various ablated versions of \textsc{Temporal-SCL} is shown.}}
    {%
        \subfigure[LSTM]{%
            \includegraphics[width=0.23\textwidth]{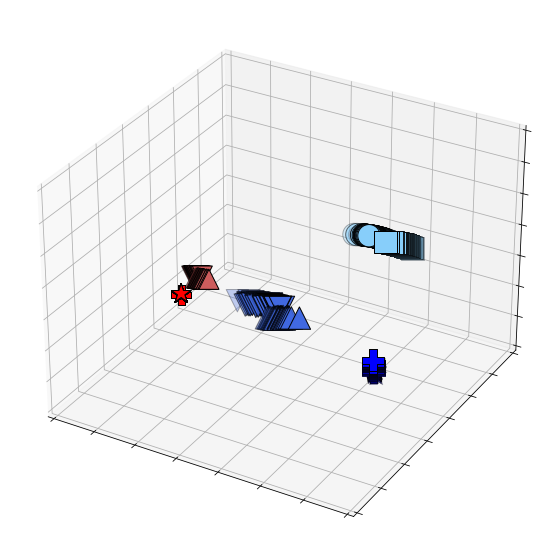}
        }
        \subfigure[RETAIN]{%
            \includegraphics[width=0.23\textwidth]{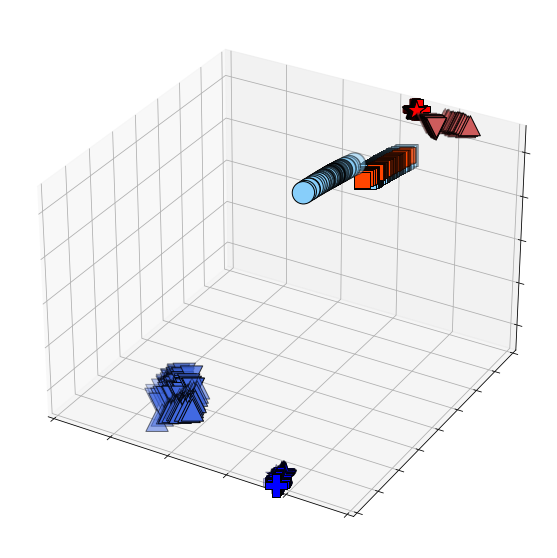}
        }
        \subfigure[DIPOLE]{%
            \includegraphics[width=0.23\textwidth]{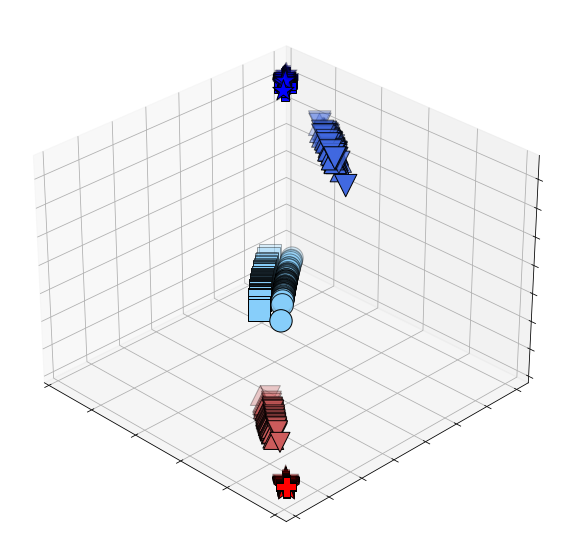}
        }
        \subfigure[Transformer]{%
            \includegraphics[width=0.23\textwidth]{Figures/Embedding_Transformer.png}
        }
        \subfigure[AC-TPC]{%
            \includegraphics[width=0.23\textwidth]{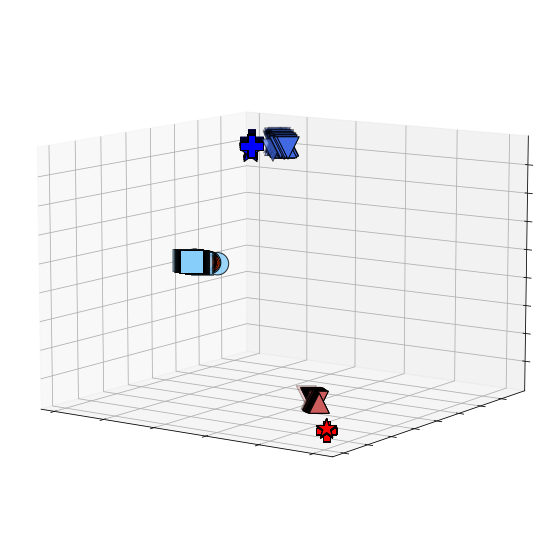}
        }
        \subfigure[T-Phenotype]{%
            \includegraphics[width=0.23\textwidth]{Figures/Embedding_TPhenotype.png}
        }
        \subfigure[SMD-SSL + SimCLR (Frozen)]{%
            \includegraphics[width=0.23\textwidth]{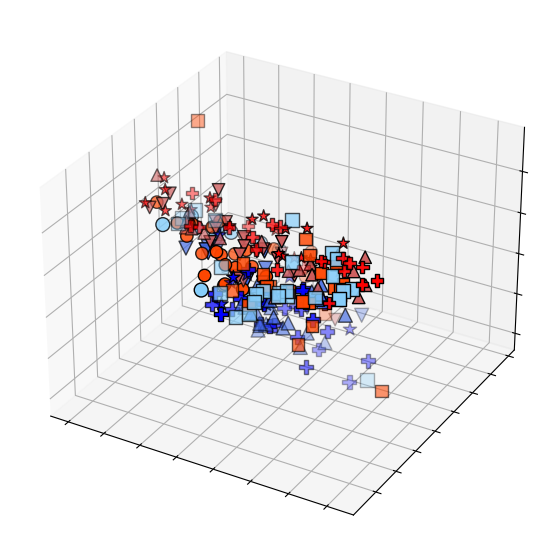}
        }
        \subfigure[SMD-SSL + VICReg (Frozen)]{%
            \includegraphics[width=0.23\textwidth]{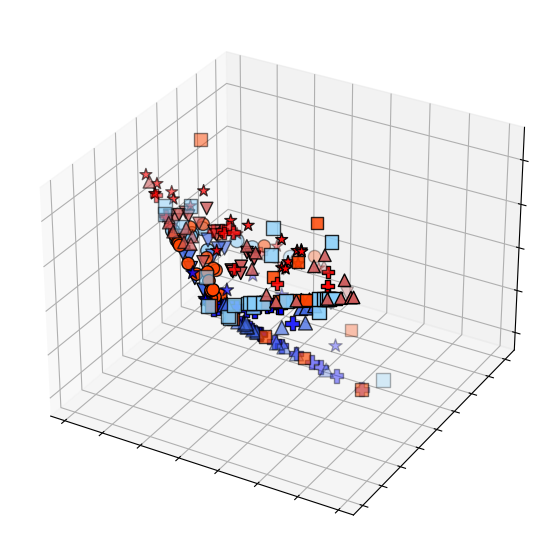}
        }
        \subfigure[SMD-SSL + SimCLR (Fine-tune)]{%
            \includegraphics[width=0.23\textwidth]{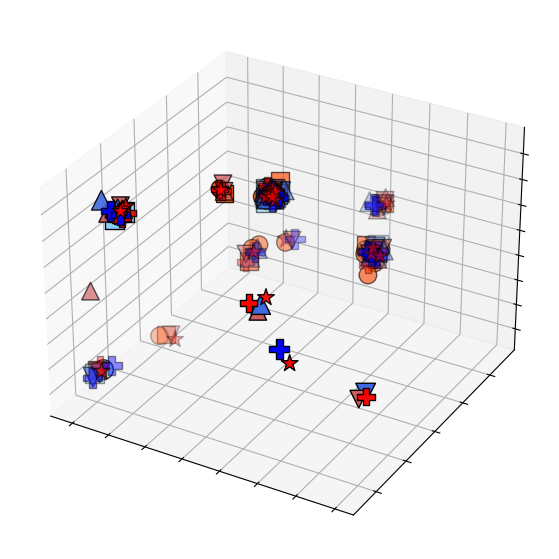}
        }
        \subfigure[SMD-SSL + VICReg (Fine-tune)]{%
            \includegraphics[width=0.23\textwidth]{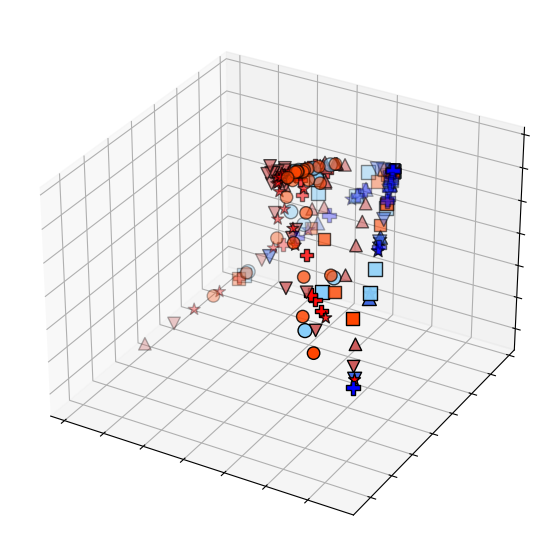}
        }
        \subfigure[\textsc{Simple-SCL}]{%
            \includegraphics[width=0.23\textwidth]{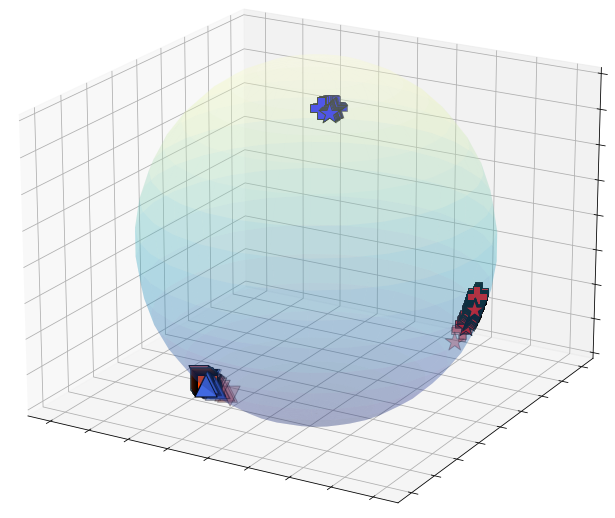}
        }
        \subfigure[\textsc{Temporal-SCL} \newline no pretraining]{%
            \includegraphics[width=0.23\textwidth]{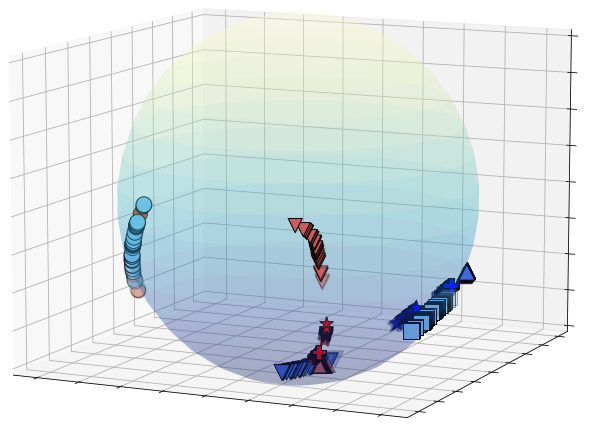}
        }
        \subfigure[\textsc{Temporal-SCL} \newline no NN pairing]{%
            \includegraphics[width=0.23\textwidth]{Figures/Embedding_NoKNN.png}
        }
        \subfigure[\textsc{Temporal-SCL}]{%
            \includegraphics[width=0.23\textwidth]{Figures/Embedding_TSCL.png}
        }
    }
\end{figure*}

As it can be seen here, our proposed model is the only model that can fully recover the structure and the 10 clusters of time steps from the input raw feature space shown in Figure~\ref{fig:toyDataset_Raw}.

\subsubsection{Visualizing Embedding Clusters for MIMIC}
\label{sec:heatmap_mimic}
We show the full heatmap of Figure~\ref{fig:heatmapMIMIC} for how the features vary across clusters in the test set of our MIMIC dataset in Figure~\ref{fig:heatmapMIMIC_full}. This heatmap is generated from the test patient embeddings of the complete \textsc{Temporal-SCL} model.
The heatmap uses the first version of our model out of the 10 experimental repeats.

\subsubsection{Visualizing Embedding Clusters for ADNI}
\label{sec:heatmap_adni}

\begin{figure*}[t!]
 \includegraphics[width=.95\linewidth]{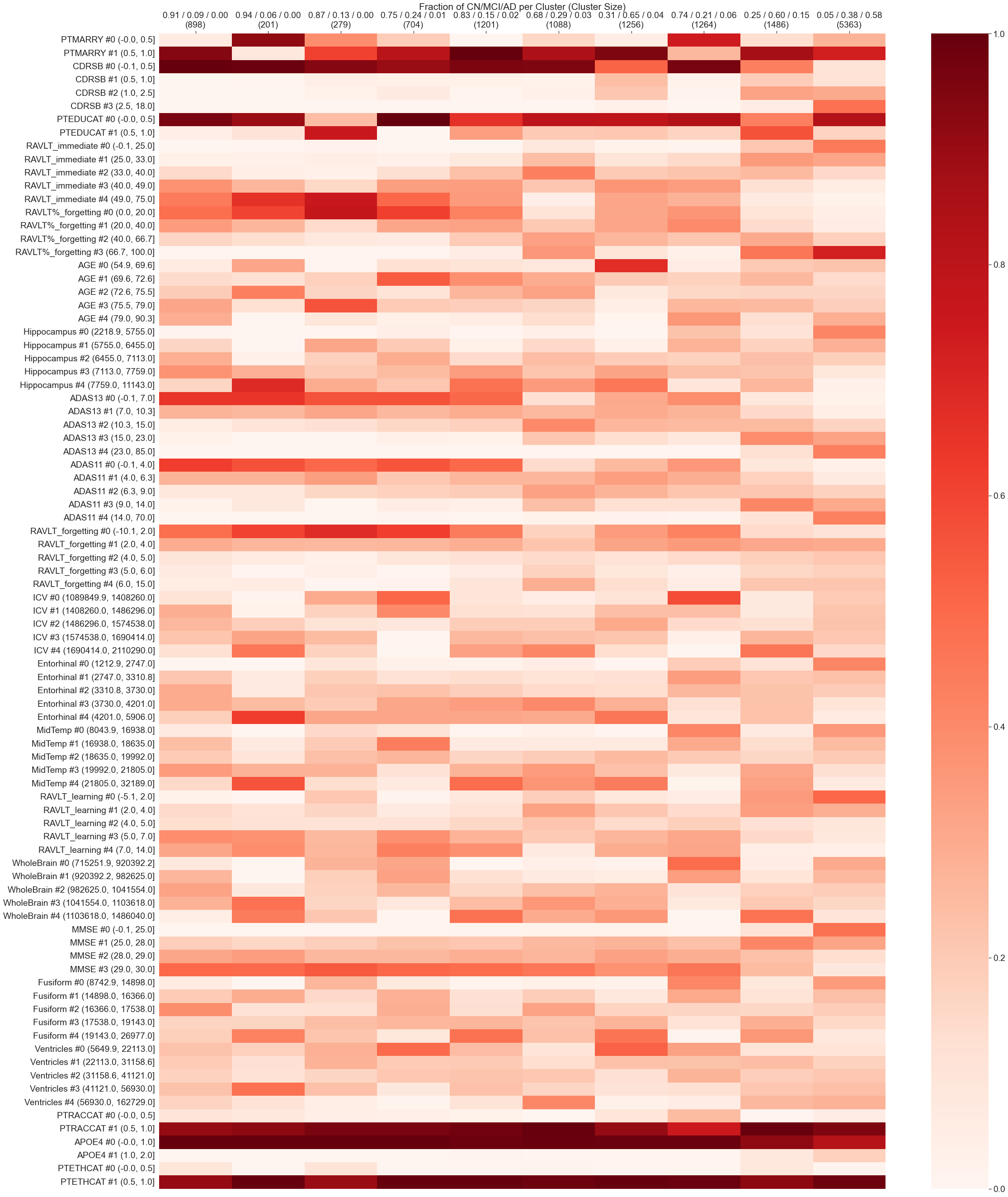}
\centering
  \caption{Heatmap showing how features (rows) vary across clusters (columns) for the ADNI dataset when using \emph{clustering on \textsc{Temporal-SCL} learned embedding space}. Heatmap intensity values can be thought of as the conditional probability of seeing a feature value (row) conditioned on being in a cluster (column); these probabilities are estimated using test set snapshots. Columns are ordered left to right in increasing fraction of test set snapshots that come from a time series that has a final outcome of Alzheimer's Disease.
 }
    \label{fig:heatmapADNI}  
\end{figure*}

To interpret each cluster for ADNI, we plot the heatmap how features (rows) vary across clusters for the test patients of ADNI in Figure \ref{fig:heatmapADNI}.
Columns are ordered (left to right) in dementia rate. 
Here we can also see that abnormal feature values that are correlated with higher risk of dementia such as irregular dementia rating scores are present in clusters containing higher proportion of AD. The heatmap uses the first version of our model out of the 10 experimental repeats.

\subsubsection{Visualizing Embedding Clusters for MIMIC from Raw Features}
\label{sec:heatmap_mimic_raw}
\begin{figure*}[t!]
\vspace{-1em}\centering
 \includegraphics[width=.9\linewidth]{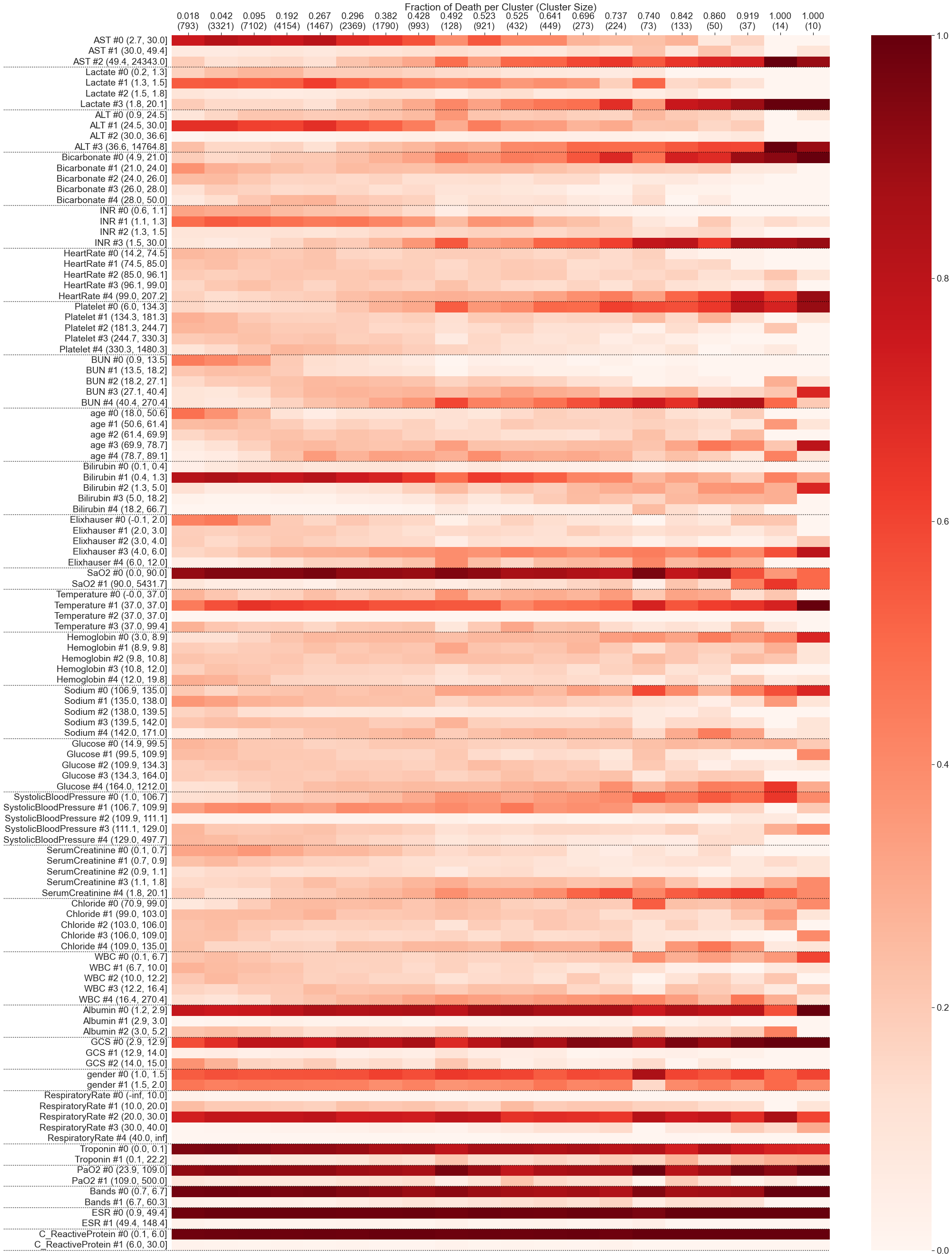}
 \vspace{-.5em}
  \caption{Heatmap showing how features (rows) vary across clusters (columns) for the sepsis cohort of the MIMIC dataset when using \emph{clustering on \textsc{Temporal-SCL} learned embedding space}. Heatmap intensity values can be thought of as the conditional probability of seeing a feature value (row) conditioned on being in a cluster (column); these probabilities are estimated using test set snapshots. Columns are ordered left to right in increasing fraction of test set snapshots that come from a time series that has a final outcome of death.
 }
    \label{fig:heatmapMIMIC_full}  
\end{figure*}

\begin{figure*}[t!]
 \includegraphics[width=.88\linewidth]{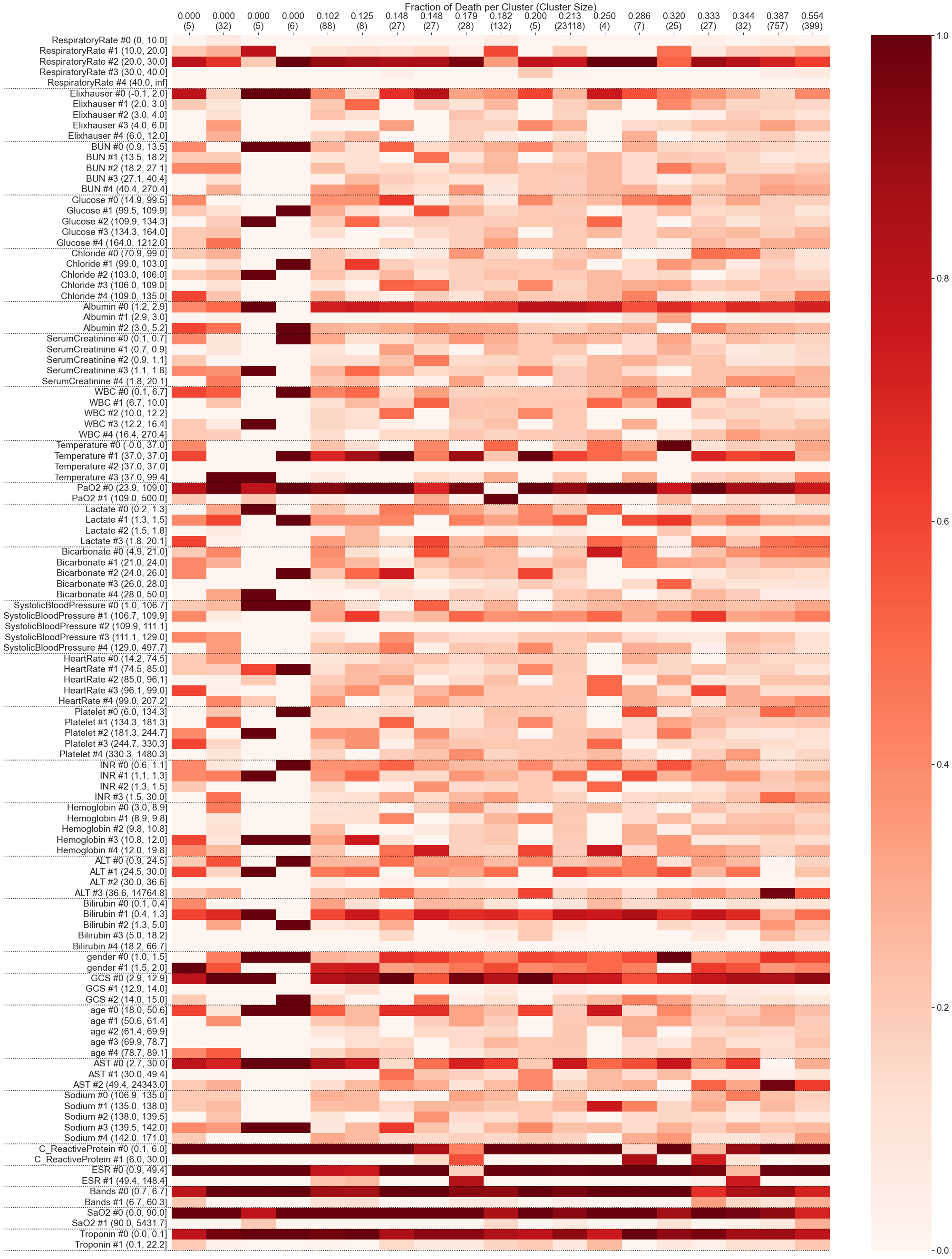}
\centering
  \caption{Heatmap showing how features (rows) vary across clusters (columns) for the MIMIC dataset when using \emph{clustering on the Raw features} instead of model embeddings. Heatmap intensity values can be thought of as the conditional probability of seeing a feature value (row) conditioned on being in a cluster (column); these probabilities are estimated using test set snapshots. Columns are ordered left to right in increasing fraction of test set snapshots that come from a time series that has a final outcome of death.
 }
    \label{fig:RawHeatmapADNI}  
\end{figure*}

We also tried our clustering-based heatmap visualization for clusters from raw feature vectors instead of \textsc{Temporal-SCL}’s embedding vectors. Using raw features results in less useful heatmaps. For example, in MIMIC as shown in Figure~\ref{fig:RawHeatmapADNI}, using raw feature vectors and 20 clusters, the resulting heatmap has high concentration in one cluster (93\% of data are in a single cluster) and the mortality rates across clusters are nearly all below 35\% (so the clusters found do not distinguish between patients with intermediate mortality probabilities, e.g., between 50\% to 100\%). Neither of these issues arise when we cluster on our model’s embedding vectors as shown in Figure~\ref{fig:heatmapMIMIC_full}.

\subsubsection{Ablation Experiments}
\label{sec:ablation}

We conduct an ablation study of excluding the three modules of our framework: (1) pre-training, (2) nearest neighbor pairing, and (3) temporal network. 
Including or withholding each of these modules results in the following ablated models respectively:
\begin{enumerate}
    \item \textsc{Simple-SCL} \\(Pre-Training: \xmark, NN Pairing: \xmark, Temporal: \xmark)
    \item \textsc{Temporal-SCL} - Pretrain Only \\ (Pre-Training: \cmark, NN Pairing: \xmark, Temporal: \xmark)
    \item \textsc{Temporal-SCL} - NN Pairing Only \\ (Pre-Training: \xmark, NN Pairing: \cmark, Temporal: \xmark)
    \item \textsc{Temporal-SCL} - Temporal Only \\ (Pre-Training: \xmark, NN Pairing: \xmark, Temporal: \cmark)
    \item \textsc{Temporal-SCL} - No Pretrain \\ (Pre-Training: \xmark, NN Pairing: \cmark, Temporal: \cmark)
    \item \textsc{Temporal-SCL} - No NN Pairing \\ (Pre-Training: \cmark, NN Pairing: \xmark, Temporal: \cmark)
    \item \textsc{Temporal-SCL} - No Temporal \\ (Pre-Training: \cmark, NN Pairing: \cmark, Temporal: \xmark)
    \item \textsc{Temporal-SCL} - Full \\ (Pre-Training: \cmark, NN Pairing: \cmark, Temporal: \cmark)
\end{enumerate}
We observe that together, all three modules play a key role in performance for both our synthetic and real-world clinical datasets experiments. 

\begin{table*}[ht!]
\caption{Ablation Study: Supervised and unsupervised performance.}
\setlength\tabcolsep{4pt}
\centering
\adjustbox{max width=\textwidth} {%
\begin{tabular}{cccccccc}
\toprule
\multirow{2}{*}[-2pt]{Dataset} &    \multirow{2}{*}[-2pt]{Model}    & \multicolumn{2}{c}{Supervised}                            & \multicolumn{4}{c}{Unsupervised} \\
                      \cmidrule(lr){3-4} \cmidrule(lr){5-8} 
                       &                                    & AUROC                      & AUPRC                    & Purity                    & NMI                       & ARI                       & SI \\ 
                       \midrule
                       \multirow{8}{*}{MIMIC}                      
                & \textsc{Simple-SCL} / \textsc{Temporal-SCL} (PT:\xbmark, NN:\xbmark, TR:\xbmark)                       & 0.753$\pm$0.005          & 0.498$\pm$0.003            & 0.773$\pm$0.005           & 0.007$\pm$0.003           & 0.003$\pm$0.002           & 0.011$\pm$0.010 \\
                & \textsc{Temporal-SCL} (PT:\cbmark, NN:\xbmark, TR:\xbmark)      & 0.752$\pm$0.002	       & 0.498$\pm$0.001	        & 0.770$\pm$0.005           & 0.006$\pm$0.003           & 0.002$\pm$0.002           & 0.011$\pm$0.001 \\ 
                & \textsc{Temporal-SCL} (PT:\xbmark, NN:\cbmark, TR:\xbmark)   & 0.770$\pm$0.002	       & 0.516$\pm$0.003            & 0.788$\pm$0.002           & 0.093$\pm$0.006           & 0.108$\pm$0.031           & 0.269$\pm$0.021 \\
                & \textsc{Temporal-SCL} (PT:\xbmark, NN:\xbmark, TR:\cbmark)      & 0.766$\pm$0.001	       & 0.508$\pm$0.002	        & 0.789$\pm$0.001           & 0.111$\pm$0.003           & 0.096$\pm$0.008           & 0.141$\pm$0.038 \\
                & \textsc{Temporal-SCL} (PT:\xbmark, NN:\cbmark, TR:\cbmark)       & 0.754$\pm$0.006          & 0.499$\pm$0.003            & 0.775$\pm$0.000           & 0.007$\pm$0.002           & 0.002$\pm$0.001           & 0.031$\pm$0.018 \\
			& \textsc{Temporal-SCL} (PT:\cbmark, NN:\xbmark, TR:\cbmark)      & 0.767$\pm$0.002          & 0.511$\pm$0.002            & 0.781$\pm$0.000           & 0.112$\pm$0.007           & {\bftab 0.150$\pm$0.034}  & 0.143$\pm$0.007 \\ 
                & \textsc{Temporal-SCL} (PT:\cbmark, NN:\cbmark, TR:\xbmark)        & 0.770$\pm$0.002	       & 0.518$\pm$0.003	        & 0.776$\pm$0.001           & 0.091$\pm$0.004           & 0.097$\pm$0.009           & 0.398$\pm$0.055 \\
                & \textsc{Temporal-SCL} (Full)              & {\bftab 0.773$\pm$0.002} &{\bftab 0.520$\pm$0.003}    & {\bftab 0.793$\pm$0.001}  & {\bftab 0.115$\pm$0.002}  & 0.128$\pm$0.014           & {\bftab 0.423$\pm$0.065} \\ 
                       \midrule
\multirow{8}{*}{ADNI}  
                & \textsc{Simple-SCL} / \textsc{Temporal-SCL} (PT:\xbmark, NN:\xbmark, TR:\xbmark)                        & 0.947$\pm$0.001          & 0.894$\pm$0.015            & 0.639$\pm$0.023           & 0.159$\pm$0.058           & 0.230$\pm$0.012           & 0.177$\pm$0.144   \\
                & \textsc{Temporal-SCL} (PT:\cbmark, NN:\xbmark, TR:\xbmark)      & 0.947$\pm$0.001	       & 0.894$\pm$0.014	        & 0.640$\pm$0.029           & 0.163$\pm$0.042           & 0.219$\pm$0.018           & 0.176$\pm$0.150 \\
                & \textsc{Temporal-SCL} (PT:\xbmark, NN:\cbmark, TR:\xbmark)   & 0.987$\pm$0.001	       & 0.899$\pm$0.011	        & 0.753$\pm$0.010           & 0.408$\pm$0.026           & 0.288$\pm$0.017           & 0.199$\pm$0.029 \\
                & \textsc{Temporal-SCL} (PT:\xbmark, NN:\xbmark, TR:\cbmark)      & 0.950$\pm$0.002	       & 0.785$\pm$0.015	        & 0.685$\pm$0.025           & 0.262$\pm$0.016           & 0.210$\pm$0.071           & 0.159$\pm$0.101 \\           
                & \textsc{Temporal-SCL} (PT:\xbmark, NN:\cbmark, TR:\cbmark)        & 0.967$\pm$0.002          & 0.900$\pm$0.017            & 0.713$\pm$0.095           & 0.275$\pm$0.120           & 0.209$\pm$0.104           & 0.149$\pm$0.080 \\ 
			& \textsc{Temporal-SCL} (PT:\cbmark, NN:\xbmark, TR:\cbmark)      & 0.951$\pm$0.002          & 0.762$\pm$0.010            & 0.749$\pm$0.009           & 0.446$\pm$0.009           & 0.334$\pm$0.015           & 0.163$\pm$0.008 \\ 
                & \textsc{Temporal-SCL} (PT:\cbmark, NN:\cbmark, TR:\xbmark)        & 0.988$\pm$0.001	       & 0.903$\pm$0.011	        & 0.753$\pm$0.011           & 0.415$\pm$0.024           & 0.300$\pm$0.042           & 0.197$\pm$0.033 \\
                & \textsc{Temporal-SCL} (Full)              & {\bftab 0.990$\pm$0.004} & {\bftab 0.936$\pm$0.014}   & {\bftab 0.755$\pm$0.023}  & {\bftab 0.452$\pm$0.031}  & {\bftab 0.399$\pm$0.010}  & {\bftab 0.259$\pm$0.031} \\
\bottomrule
\end{tabular}}

\vspace{.5ex}
{\scriptsize {\bftab{PT}}: Pre-Training, {\bftab{NN}}: Nearest Neighbor pairing, {\bftab{TR}}: Temporal Regularization\\}
\label{tab:ablation_results}
\end{table*}

\textbf{Synthetic dataset.} 
In Table~\ref{tab:toyDataset_ratios}, examining the ablated models reveals the distinct contributions of each module in our synthetic experiments. Specifically, the observed decrease in Silhouette Index (SI) when NN pairing is absent underscores its pivotal role in maintaining feature similarity, which is instrumental in ground truth recovery. Conversely, the model lacking pre-training suggests that while pre-training facilitates initial model configuration, the presence of NN pairing and the temporal network can partly mitigate its absence.

These results highlight the NN pairing module's significance in organizing the embedding space, a notion further evidenced by the relatively higher SI in the NN Pairing Only ablation (3) compared to other single-module ablations.

The temporal network's role, while not markedly influencing performance metrics in isolation, proves vital for the temporal structure within the embedding space. This is contrasted with the \textsc{Simple-SCL} model, where the simultaneous absence of NN pairing and the temporal network leads to the lowest performance metrics, emphasizing the necessity of their collective functionality.

The full \textsc{Temporal-SCL} configuration exhibits superior SI and Cluster Prediction Accuracy, demonstrating the synergistic effect of the three modules in both the recovery of input structure and the enhancement of downstream task prediction.

In Figure~\ref{fig:toyDataset_result_complete}, we include ablations of the most simple model (1) and ablations of single module omitted (6), (7), alongside the fully-equipped model (8). The learned embedding space from these models illustrates high prediction accuracy, as evidenced by a hyperplane nearly perfectly segregating the red data points from the blue data points. However, the figure also showcases the collective capacity of the modules to recover the known ground-truth embedding vectors. Notably, only the complete model represented in Figure~\ref{fig:synth-tscl} achieves an accurate reconstruction of the embedding structure, evidenced by distinct, correctly ordered clusters for each timestep.

\vspace{.2em}
\noindent
\textbf{Clinical datasets.} We present an ablation study of our model to see how each modification contributes to the performance.
Our full model includes pre-training with \textsc{Simple-SCL}, enhancing the Temporal Network $h$, and using ``labels$+$feature-similarity'' for finding the nearest neighbor pairs. 
We train the extra 7 additional ablated models in this section on both our MIMIC and ADNI datasets.
Note that to discover discrete latent states that could be of interest, we cluster on the learned hyperspherical embeddings (the $\mathbf{z}_i^{(\ell)}$ variables)  using a complete linkage Agglomerative Hierarchical Clustering \citep{murtagh2012algorithms}. We describe our clustering approach in Appendix~\ref{sec:cluster}. The unsupervised metrics presented here are evaluated for the same number of clusters in each ablation.

\begin{table*}[ht!]
\caption{Future Prediction on ADNI data test set accuracy (mean $\pm$ std.~dev.~across 10 experiments).}
\centering
\setlength\tabcolsep{8pt}
\scriptsize
\begin{tabular}{ccccc} 
\toprule
		\multirow{2}{*}[-2pt]{Model}                                       & \multicolumn{2}{c}{6 months in the future}                          & \multicolumn{2}{c}{1 year in the future}                                                   \\ \cmidrule(lr){2-3} \cmidrule(lr){4-5} 
                                                               & AUROC                    & AUPRC                   & AUROC                   & AUPRC \\ 
                       \midrule  
                        Logistic Regression                    & 0.825$\pm$0.007          & 0.659$\pm$0.008         & 0.767$\pm$0.006          & 0.620$\pm$0.012 \\
					  LSTM                                   & \text{0.880$\pm$0.003}   & 0.789$\pm$0.005         & 0.856$\pm$0.004          & 0.736$\pm$0.008 \\
                        RETAIN                                 & \text{0.882$\pm$0.010}   & 0.793$\pm$0.016         & \text{0.867$\pm$0.013}   & 0.740$\pm$0.017 \\
                        DIPOLE                                 & \text{0.940$\pm$0.006}   & 0.864$\pm$0.009         & \text{0.918$\pm$0.008}   & 0.813$\pm$0.010 \\
                        AC-TPC                                 & 0.823$\pm$0.011          & 0.702$\pm$0.016         & 0.804$\pm$0.014          & 0.679$\pm$0.020 \\
                        Transformer:BERT                       & \text{0.947$\pm$0.005}   & 0.880$\pm$0.010         & \text{0.939$\pm$0.007}   & \text{0.846$\pm$0.011} \\
                        \cmidrule(lr){1-5}
                        \textsc{Simple-SCL}                    & 0.922$\pm$0.005          & 0.794$\pm$0.016         & 0.899$\pm$0.003          & 0.764$\pm$0.015    \\
                        \textsc{Temporal-SCL} (PT:\xbmark, NN:\cbmark, TR:\cbmark)    & 0.950$\pm$0.002          & {0.878$\pm$0.014}       & 0.930$\pm$0.002          & {0.807$\pm$0.018}  \\ 
					 \textsc{Temporal-SCL} (PT:\cbmark, NN:\xbmark, TR:\cbmark)    & 0.948$\pm$0.002          & {0.842$\pm$0.012}       & 0.929$\pm$0.002          & {0.800$\pm$0.011}  \\ 
                       \textsc{Temporal-SCL} (Full)            & {\bftab 0.985$\pm$0.002} &{\bftab 0.907$\pm$0.012} & {\bftab 0.970$\pm$0.003} & {\bftab 0.878$\pm$0.015}  \\ 
\bottomrule
\end{tabular}

\vspace{.5ex}
{\scriptsize {\bftab{PT}}: Pre-Training, {\bftab{NN}}: Nearest Neighbor pairing, {\bftab{TR}}: Temporal Regularization\\}
\label{tab:results_future_adni}
\end{table*}

Firstly, our empirical findings show that for the models without pre-training and without the temporal network $h$ (\textsc{Simple-SCL}), we see a clear performance drop for all supervised and unsupervised metrics which highlights the importance of inclusion of these modules in our model training.
We also calculated the Silhouette Index (SI) of the ablated models in the last column of Table~\ref{tab:ablation_results}. As it can be seen from the ablation results, with respect to the supervised prediction performance (AUROC, AUPRC), the two models with and without NN pairing ((5), (8)) perform similarly. However, the main gain of using “labels+feature similarity” comes in the unsupervised prediction performance (Purity, NMI, ARI, SI). This is especially apparent in the Silhouette Index score (a measure of how similar an object is to its own cluster compared to other clusters) where we see the greatest boost in performance when using “labels+feature similarity” (full model (8)) instead of “labels only” (ablated model (5)) where it shows how our model moves away from just stratifying risk (which is what the supervised metrics are measuring) to additionally being capable of identifying homogeneous disease phenotypes.
These experiments together, underscore the significance of having the different building blocks of our model for achieving the highest performance gain in our experiments.

Additionally, the ablation results shown in the bottom portion of Table~\ref{tab:results} reveal that the highest performance gain with respect to the supervised prediction performance (AUROC, AUPRC) is achieved by the full \textsc{Temporal-SCL}.
In our MIMIC experiment, the ablated model without only the nearest neighbor pairing achieves the closest performance to the full model, while in ADNI, where the prediction task seems to be easier (as evident by the high prediction scores), the ablated model without pre-training and the ablated model without the temporal regularization of our temporal network are the closests to the full model with respect to the prediction performance.
These experiments again underscore the significance of the different building blocks of our model to achieve the greatest performance gain in our experiments within MIMIC and ADNI datasets.

\subsubsection{Additional Experiments on ADNI: Predicting Future Cognitive Impairment Status}
\label{sec:future_adni}
In our experiments presented in \ref{sec:ClinicalExperiments}, we followed the same experimental paradigm of predicting the current time step label as conducted by \cite{lee2020temporal}. 
In this section however we will look at an additional prediction task for predicting the cognitive impairment 6 months and 1 year in the future. 
The results are presented in Table~\ref{tab:results_future_adni} below. 
We can see that for this prediction task again our model outperforms other baselines.

\end{document}